\documentclass[11pt]{article}

\usepackage[margin=1in]{geometry}
\usepackage{times}
\usepackage{latexsym}
\usepackage[T1]{fontenc}
\usepackage[utf8]{inputenc}
\usepackage{microtype}
\usepackage{inconsolata}
\usepackage{amsmath,amsfonts}
\usepackage{graphicx}
\usepackage{booktabs}
\usepackage{multirow}
\usepackage{xcolor}
\usepackage{listings}
\usepackage{subcaption}
\usepackage{natbib}
\usepackage{hyperref}

\title{PokerSkill: LLMs Can Play Expert-Level Poker without Training or Solvers}

\author{Boning Li\thanks{IIIS, Tsinghua University. Email: li-bn22@mails.tsinghua.edu.cn}
\and Baoxiang Wang\thanks{The Chinese University of Hong Kong, Shenzhen.}
\and Longbo Huang\thanks{IIIS, Tsinghua University. Email: longbohuang@tsinghua.edu.cn. Corresponding author.}}

\date{\today}

\begin{document}
\maketitle

\begin{abstract}
Poker is a landmark challenge for artificial intelligence. The dominant approach relies on equilibrium solvers built on counterfactual regret minimization, requiring millions of core-hours of training. Large Language Models (LLMs) possess extensive poker knowledge but perform far below solver-based agents when asked to play directly. Traditional rule-based poker agents are interpretable and training-free, but their strategic ceiling remains far below equilibrium play. 
We introduce \textbf{PokerSkill}, a training-free and solver-free framework that bridges this gap by using detailed rule-based poker skills as a structured action-grounding interface for LLMs. A deterministic context engine analyzes the current state and retrieves only the relevant fragments from a layered skill library, which is entirely designed by human poker experts, constraining the LLM's choice to reasonable actions. Against GTOWizard, a state-of-the-art GTO benchmark, GPT-5.5 XHigh with PokerSkill achieves $-57 \pm 21$ mbb/hand, Claude Opus 4.6 achieves $-80 \pm 29$ mbb/hand and Claude Opus 4.7 achieves $-87\pm 64$ mbb/hand, reducing losses by 49--61\% compared to default-prompt baselines and outperforming the strong bot Slumbot. Our key finding is that rule-based skills alone do not constitute a strong strategy, and LLMs alone cannot play well, but their combination yields an agent that requires neither training nor solver access yet competes with systems built on millions of core-hours of computation. To our knowledge, this is the first demonstration of an LLM achieving competitive performance in a complex imperfect-information game without game-specific training or solver queries. Code is available at \url{https://github.com/lbn187/PokerSkill}.

\end{abstract}

\section{Introduction}
\label{sec:intro}

\begin{figure}[t]
    \centering
    \includegraphics[width=0.95\textwidth]{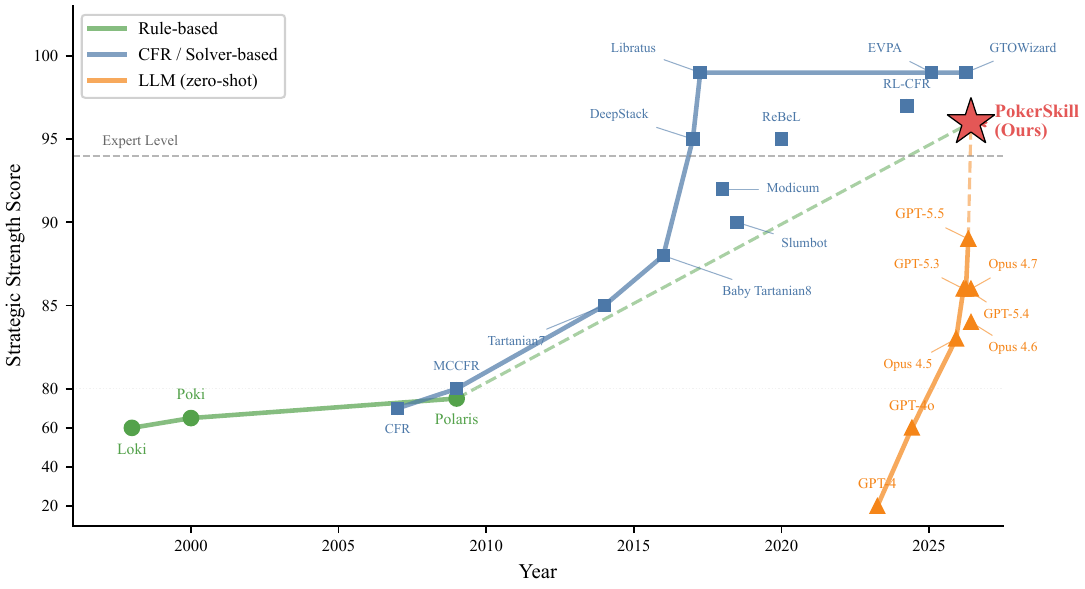}
    \caption{Evolution of poker AI systems across three paradigms. Rule-based agents (green) offer interpretability but limited strategic depth. CFR/Solver-based systems (blue) achieve near-equilibrium play through massive computation. LLM-based agents (orange) possess strategic knowledge but struggle to apply it effectively. PokerSkill (red star) bridges the gap by using rule-based skills as an action-grounding interface for LLMs, achieving competitive performance without training or solver access.}
    \label{fig:poker-ai-evolution}
\end{figure}

Poker, particularly Heads-Up No-Limit Texas Hold'em (HUNL), is among the most challenging domains in artificial intelligence \citep{billings2002challenge}. Its game tree exceeds $10^{164}$ nodes \citep{johanson2013measuring}, requiring agents to handle imperfect information, stochastic outcomes, deception, and multi-street planning within a single framework. Solving poker has driven fundamental advances in decision-making under uncertainty and remains an active frontier for AI research \citep{bowling2015heads,moravcik2017deepstack,brown2018superhuman,brown2019superhuman}. Figure~\ref{fig:poker-ai-evolution} illustrates the evolution of poker AI across three paradigms.

The most successful approaches to HUNL rely on equilibrium solvers built on Counterfactual Regret Minimization (CFR) and its variants~\citep{zinkevich2007regret,brown2019solving,li2026parallel}. These methods iteratively traverse the game tree to converge toward Nash equilibrium strategies. However, the computational cost is immense: Libratus required over 15 million core-hours of computation to achieve superhuman performance \citep{brown2018superhuman}. Even modern solvers demand dedicated GPU clusters and months of training time \citep{brown2020combining}. This computational barrier restricts equilibrium-based approaches to poker research.

A second tradition in poker AI is rule-based expert systems~\citep{billings1998opponent,davidson2000improved,billings2002challenge}. These agents encode human poker knowledge as hand-crafted heuristics for betting, calling, and folding. They are interpretable, require no training, and can be deployed instantly. However, their strategic ceiling is far below equilibrium play: rigid rules cannot represent the complex conditional strategies needed for no-limit variants, and they lack the contextual reasoning to adapt to novel situations. The gap between rule-based agents and solver-based systems has historically been hundreds of mbb/hand~\citep{rubin2011computer}. This raises a natural question: is the weakness of rule-based approaches inherent to the paradigm, or does it stem from the limited reasoning capacity of the rule executor?

Large Language Models (LLMs) present a fundamentally different paradigm. Trained on corpora that include poker strategy books, expert commentary, solver analysis outputs, and forum discussions, frontier LLMs have absorbed substantial strategic knowledge. In principle, this knowledge should enable competent play. In practice, however, LLMs perform catastrophically when asked to play HUNL directly: the GTOWizard benchmark~\citep{gtowizardbenchmark2026} reports that Claude Opus 4.6 loses $-204 \pm 44$ mbb/hand, GPT-5.4 loses $-178 \pm 37$ mbb/hand, and GPT-5.3 loses $-160 \pm 30$ mbb/hand under default prompting. These results indicate that even with the most advanced LLM, it is impossible to play poker well without additional guidance.

Why do models that can fluently explain pot odds, minimum defense frequency, and polarized ranges fail so badly at the table? We hypothesize that a core obstacle is a \emph{decision-binding problem}. At any given node, multiple strategic concepts may simultaneously relevant: board texture suggests one action, pot odds another, the opponent's line a third. The model must not merely recall these concepts but select which one should govern this specific decision. Under standard prompting, this multi-factor arbitration happens implicitly, and the model frequently binds the wrong principle to the node. A hand that warrants aggression on a dry board becomes a passive check on a wet board; ace-high is a bluff-catcher in one line and worthless in another. The model knows the difference in principle but may fail to apply it in the moment.

We introduce \textbf{PokerSkill}, a framework designed to \emph{unearth} latent poker skills from general LLMs through structured prompt guidance, requiring zero solver queries at inference time and zero offline learning. The key insight is that the strategic knowledge already exists in these models but needs an external activation signal to be correctly deployed at each decision node. PokerSkill operates as follows: at each action point, a deterministic context engine analyzes the current state and produces compact labels describing board texture, hand class, action history, position, stack-to-pot ratio, and cumulative betting pressure. These labels drive selective retrieval from a layered skill library that is entirely designed by human poker experts among the authors, who are experienced players that have spent years studying GTO theory and competing at high stakes. The library encodes detailed, reliable guidance covering $\sim$60 action-line scenarios, 23 hand classes, and 46 bet-size pressure thresholds, ensuring the LLM receives only the relevant expert knowledge for the current decision. An attack/defense budget system tracks cumulative pressure across streets and constrains the action space to strategically viable options. The LLM then exercises judgment within this bounded, context-specific guide.

Our evaluation uses GTOWizard \citep{gtowizardbenchmark2026}, which provides AIVAT variance reduction \citep{burch2018aivat} and has beaten Slumbot \citep{jackson2013slumbot} (the SoTA open-source poker bot before 2025) by $194 \pm 41$ mbb/hand over $150{,}000$ hands. With PokerSkill, GPT-5.5 XHigh achieves $-57 \pm 21$ mbb/hand against GTOWizard, Claude Opus 4.6 achieves $-80 \pm 29$ mbb/hand, and Claude Opus 4.7 achieves $-87 \pm 64$ mbb/hand. Compared to their default-prompt baselines ($-132 \pm 25$, $-204 \pm 44$, and $-170 \pm 28$ mbb/hand respectively), PokerSkill reduces losses by 49--61\%. All three PokerSkill agents achieve lower loss rates against GTOWizard than the strong bot Slumbot, establishing that pure LLM agents can reach a competitive poker play level, without any game-tree traversal,  iterative training or solver queries.

Our contributions are:
\begin{itemize}
    \item We present PokerSkill, the strongest LLM poker agent reported to date, achieving competitive performance with historical solver champion-level systems in full HUNL without solver queries at inference time or offline learning. To our knowledge, this is the first demonstration of an LLM reaching this level of play in a complex imperfect-information game.
    \item We demonstrate that detailed rule-based poker skills, which are individually insufficient for strong play, can serve as an effective action-grounding interface for LLMs. This yields a training-free, solver-free HUNL agent that improves multiple frontier LLMs (49--61\% loss reduction), showing that poker expert skills improves the poker ability for LLMs significantly.
    \item Unlike the previously difficult to understand and reproduce HUNL poker AI, PokerSkill provides a poker agent that are easy to understand, reproduce, and improve. We release the poker agent as open source, making this the first open and reproducible strong poker system. Any researcher with access to a frontier LLM API can reproduce our results, and the framework improves automatically as base models advance.
    
\end{itemize}


\section{Related Work}
\label{sec:related}

\paragraph{Poker solvers and game-theoretic agents.}
CFR and its variants~\citep{zinkevich2007regret,lanctot2009monte,brown2019solving,farina2021faster} are the foundation for solving large imperfect-information games. Superhuman HUNL systems combine CFR with abstraction~\citep{ganzfried2014potential,li2024rlcfr,li2025evpa,li2026weva}, depth-limited solving~\citep{brown2018depth,brown2020combining}, and subgame re-solving~\citep{brown2017safe}: DeepStack~\citep{moravcik2017deepstack} and Libratus~\citep{brown2018superhuman} both require millions of core-hours. GTOWizard~\citep{gtowizardbenchmark2026} represents the current state-of-the-art benchmark, beating the strong bot  Slumbot~\citep{jackson2013slumbot} by $194 \pm 41$ mbb/hand.

\paragraph{LLM poker agents.} LLMs can discuss poker strategy but struggle to play reliably. Gupta~\citep{gupta23} found ChatGPT and GPT-4 deviate from GTO strategies. Suspicion-Agent~\citep{guo2024suspicionagent} uses theory-of-mind prompting but only in small Leduc Hold'em. PokerBench~\citep{zhuang2025pokerbench} evaluates LLM poker reasoning; even fine-tuned models underperform solvers. ToolPoker~\citep{lin2026how} queries external solvers for GTO actions in Limit Hold'em but relies on solver access. All these approaches either restrict the game variant, use fine-tuning/RL, or depend on solver queries. PokerSkill targets full HUNL with a pure LLM agent requiring none of these.

\paragraph{Rule-based poker agents.}
The earliest computer poker programs relied on hand-crafted heuristics and expert systems. Loki~\citep{billings1998opponent} and its successor Poki~\citep{davidson2000improved} combined rule-based betting logic with opponent modeling. These systems demonstrated that expert knowledge could produce reasonable play in poker. Subsequent work explored game-theoretic approximations within rule-based frameworks~\citep{billings2003approximating}, and the Polaris system~\citep{bowling2009polaris} combined rule-based components with equilibrium computation. Despite their interpretability and zero training cost, rule-based agents hit a strategic ceiling well below equilibrium play \citep{rubin2011computer}. PokerSkill revisits this paradigm: detailed rule-based poker skills alone do not constitute a strong strategy, but when used as a \emph{structured action-grounding interface} for modern LLMs, which supply the contextual reasoning that rigid rule systems lack, they enable competitive HUNL performance without any training or solver access.

\paragraph{LLM agents in strategic settings.}
Cicero~\citep{meta2022human} achieved human-level Diplomacy by combining language models with strategic planning. Voyager~\citep{wang2023voyager} uses LLMs with a skill library for open-ended embodied tasks, which represents an architectural parallel to PokerSkill's layered retrieval. In social deduction games, LLM agents exhibit emergent strategic behaviors but remain suboptimal~\citep{bailis2024werewolf}. A common finding is that raw LLM capabilities require structured scaffolding for effective situated action~\citep{xi2025rise}. PokerSkill shows that \emph{domain-specific context binding}, rather than general reasoning chains, is the key bottleneck in imperfect-information sequential decisions.

\paragraph{Retrieval-augmented and tool-augmented LLMs.}
RAG augments LLMs with external knowledge at inference time~\citep{lewis2020retrieval}, and Toolformer~\citep{schick2023toolformer} teaches models to invoke tools. PokerSkill shares the principle of injecting task-relevant information at decision time. However, unlike RAG, PokerSkill's retrieval is \emph{deterministic and rule-based}: the context engine computes exact game-state features and selects prompt fragments by logical condition, reflecting the structured nature of the game state.

\paragraph{Structured prompting.}
Chain-of-thought~\citep{wei2022chain}, Tree of Thoughts~\citep{yao2023tree}, ReAct~\citep{yao2023react}, and Program-of-Thoughts~\citep{chen2023program} improve LLM performance through structured interfaces without parameter updates. However, these general-purpose scaffolds do not solve the decision-binding problem: they help the model reason more carefully but cannot tell it \emph{which} concept should govern a specific game node. PokerSkill's contribution is precisely this domain-specific binding mechanism. Broader evaluations confirm systematic LLM failures in game-theoretic settings~\citep{huang2024far,hua2024game}; PokerSkill addresses these through state-dependent context analysis. 

\section{Unearthing Poker Skills from Language Models}
\label{sec:method}

Frontier LLMs are trained on vast corpora that include poker strategy books, forum discussions, solver analyses, and expert commentary. This training exposes models to high-quality poker reasoning, yet zero-shot LLM agents perform far below expert level. We hypothesize that a major obstacle is the combination of state-grounding failures, strategic binding errors, and action-validity issues---much as a student who has read every textbook may still freeze on an exam without structured problem-solving habits. PokerSkill addresses these failures through structured prompt guidance that combines expert state abstraction with LLM judgment at each decision point.

\begin{figure}[!h]
    \centering
    \includegraphics[width=\textwidth]{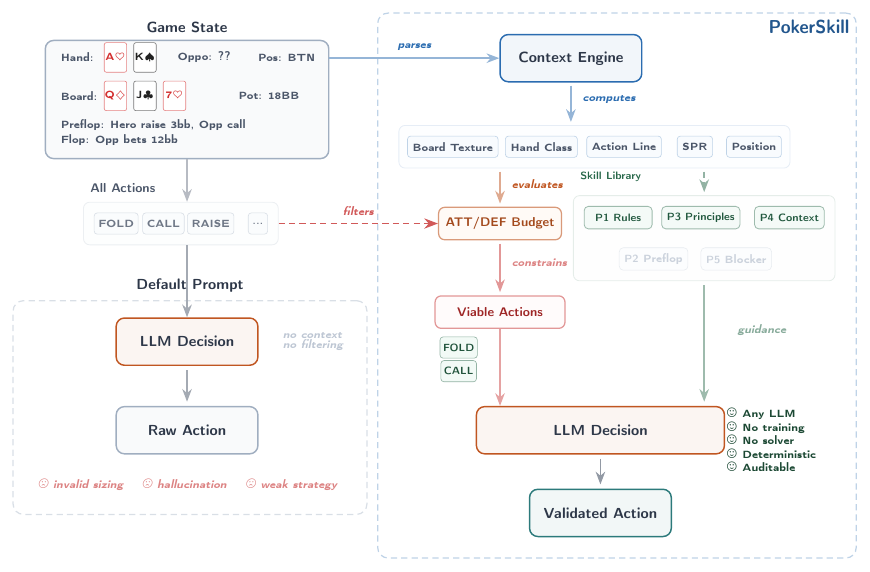}
    \caption{Overview of PokerSkill versus the default prompting baseline. \textbf{Left}: a bare LLM receives the game state and all legal actions without context, producing raw outputs prone to invalid sizing, hallucination, and weak strategy. \textbf{Right}: PokerSkill interposes a deterministic Context Engine that extracts structured features (board texture, hand class, action line, SPR, position), an ATT/DEF Budget mechanism that filters the action space into viable options, and a layered Skill Library that provides situation-specific strategic guidance. The LLM decides within these bounds, and a validator ensures the final output is a legal action.}
    \label{fig:overview}
\end{figure}

\subsection{The Decision-Binding Problem}
\label{sec:binding-problem}

The gap between knowing poker concepts and applying them correctly is what we call the \emph{decision-binding problem}. A frontier LLM can explain pot odds, blockers, minimum defense frequency, and polarized ranges when asked in isolation. The failure occurs when a live decision requires the model to simultaneously evaluate relative hand strength, board texture, betting history, position, stack depth, and legal sizing, then select the single concept that should govern the action. For example \citep{gtowizardbenchmark2026}, Claude Opus 4.6, holding 4$\heartsuit$7$\spadesuit$ on a board of 5$\heartsuit$4$\clubsuit$3$\heartsuit$4$\spadesuit$3$\spadesuit$, states: ``I have QKo on a 5d4c3h4s board---complete air with no pair.'' It hallucinates entirely different hole cards while holding trip fours. These are not reasoning errors in the usual sense; the model fails at the more basic task of correctly reading the game state.


\subsection{Framework Overview}
\label{sec:core-idea}

PokerSkill addresses decision-binding through three stages (Figure~\ref{fig:overview}): (i) \textbf{Context Engine}: Deterministic rules analyze the current state including cards, board texture, action history, position, stack-to-pot ratio (SPR), and cumulative betting pressure, producing compact labels. (ii) \textbf{Skill Retrieval}: Context labels select specific prompt fragments from a layered skill library, rather than inserting the entire library into every prompt. (iii) \textbf{Bounded Decision}: The LLM chooses among context-compatible actions within validated legal bounds.

The design philosophy mirrors how expert human players think: they do not re-derive game theory from scratch at each decision. Instead, they (1) read the situation (board texture, position, opponent's line), (2) recall the relevant strategic principle for that situation, and (3) choose among a small set of reasonable actions. PokerSkill externalizes this cognitive pipeline as a deterministic scaffold around the LLM. Importantly, ``solver-free'' means that no equilibrium solver is invoked at inference time. The skill library is informed by GTO theory and solver study---just as human professionals learn from solver analysis before sitting at the table---but the deployed system performs no game-tree traversal or iterative computation.

Critically, the entire skill library, including all prompt fragments, budget tables, context labels, and action constraints, is designed by human poker experts among the authors. The authors include one or more experienced poker players who have spent years studying GTO theory, coaching other players, and competing at high stakes. Every budget value, every hand-class boundary, and every action-line scenario reflects hard-won expert intuition about how poker should be played. This expert-driven design ensures that the guidance is detailed (covering $\sim$60 action-line scenarios, 23 hand classes, and 46 bet-size pressure thresholds), reliable (grounded in established poker theory rather than ad-hoc heuristics), and robust (the same tables generalize across three different LLMs without modification). In essence, PokerSkill distills expert poker knowledge into a structured format that LLMs can reliably execute.

\subsection{Context-Conditioned Skill Retrieval}
\label{sec:retrieval}

The skill library is organized into five layers, each activated by different context conditions:

\begin{itemize}
    \item \textbf{P1} (always active): Game rules, legal actions, output format constraints.
    \item \textbf{P2} (range table): Range guidance for the current preflop scenario.
    \item \textbf{P3} (general principles): Stable principles including value/bluff separation, positional strategy, pot control, and sizing discipline.
    \item \textbf{P4} (context-specific): Strategy keyed by board texture, hand class, action line, and role.
    \item \textbf{P5} (blocker): Blocker-aware bluffing and bluff-catching guidance for river decisions.
\end{itemize}

A preflop decision retrieves P1 + the relevant P2 entry. A postflop decision retrieves P1 + P3 + P4 entries matching the board texture and hand class. A river decision additionally retrieves P5. This selective retrieval prevents two failure modes: (1) vague general advice that does not resolve the specific decision, and (2) overwhelming detail that forces the model to arbitrate among irrelevant heuristics. We now describe each context module in detail, explaining both its poker-strategic motivation and its role in helping the LLM.

\paragraph{Preflop range tables (P2).}
In HUNL, preflop decisions are the foundation of all subsequent play. Human players spend months memorizing which hands to raise, call, or fold in each preflop scenario. The preflop game tree can be decomposed into a finite set of well-studied scenarios (e.g., button open, big blind facing a 3-bet). For each scenario, game-theoretic solvers produce precise frequency tables specifying the equilibrium action for every hand.

We encode 12 preflop scenarios covering the complete HUNL preflop decision tree at 200BB depth, from initial open/limp decisions through 5-bet and all-in confrontations. Each scenario provides frequency-annotated ranges authored by the poker experts among the authors based on their knowledge of equilibrium play. At decision time, the context engine identifies which scenario applies and injects \emph{only that single entry} into the prompt. This module is well-suited for LLMs because preflop decisions are essentially a lookup problem over discrete, enumerable factors. Without precise frequency anchors, models may systematically deviate from equilibrium. The range tables provide the calibration signal that the model's parametric knowledge lacks.

\paragraph{Stable postflop principles (P3).}
P3 encodes the foundational principles that every competent player internalizes but that LLMs frequently violate under default prompting. These include: (1) \emph{value/bluff separation}, the requirement that bets should be polarized into strong hands betting for value and weak hands bluffing, with medium-strength hands preferring to check; (2) \emph{positional strategy}, adapting aggression and defense thresholds based on whether the player acts first or last; (3) \emph{pot control}, the discipline to avoid inflating the pot with vulnerable holdings that cannot withstand a raise; and (4) \emph{sizing discipline}, matching bet sizes to hand polarity and board texture rather than defaulting to arbitrary amounts.

These principles are ``stable'' in the sense that they apply across all postflop situations regardless of specific hand class or action line. They serve as a baseline error-prevention layer: without P3, LLMs routinely commit elementary mistakes such as betting medium-strength hands into large pots, using identical strategies in and out of position, or choosing bet sizes that do not correspond to any coherent range construction. P3 is included in every postflop prompt, providing the strategic floor that context-specific modules (P4, P5) build upon.

\paragraph{Action-line scenario recognition (P4 action lines).}
In poker, the sequence of actions across streets carries critical strategic information. An experienced player instantly recognizes patterns like ``opponent bet flop, bet turn, now checks river'' as a specific \emph{line} that constrains the opponent's likely holdings. This pattern recognition---determining who is the aggressor, who is defending, and what the action sequence implies about range composition---is fundamental to expert play.

The action-line analyzer turns betting history into a scenario ID and a cumulative pressure estimate. Scenario IDs distinguish whether the hero is the aggressor or defender, whether the opponent has checked or bet, and whether the line is a c-bet, probe, donk, check-raise, or others. This scenario ID selects the relevant P4/P5 Blocker text and changes viable-action computation.

We define approximately 60 action-line scenarios across flop (9), turn (17), and river (27+), each encoding a distinct strategic situation. For example, scenario \texttt{R-AA1} (``we bet flop and turn, opponent called both, checks river'') implies the opponent's range is \emph{capped} (strong hands would have raised earlier), enabling polarized river bets. Scenario \texttt{T-D2} (``opponent bet both flop and turn'') signals sustained aggression, requiring tighter defense thresholds.

Weighted pressure is computed from bet size relative to the pot. Small bets consume little budget; pot-sized or overbet actions consume much more. The monotonic principle is:
\begin{align*}
\text{larger bet fraction}
&\Rightarrow \text{larger pressure weight} \\
&\Rightarrow \text{less remaining ATT/DEF}.
\end{align*}
The remaining budget after previous actions determines whether the hero still has permission to attack or defend.

For LLMs, action-line recognition solves a critical context-window problem. As hands progress across multiple streets, the raw action history grows long and complex. By compressing the action history into a semantic label with an attached strategic summary, we offload multi-step pattern recognition to deterministic code and present the LLM with the \emph{conclusion} rather than the raw evidence.




\paragraph{Board texture classification (P4 board texture).}
The community cards fundamentally determine which hands are strong and which strategies are appropriate. A ``dry'' board (e.g., K$\spadesuit$7$\diamondsuit$2$\clubsuit$) favors the preflop aggressor and supports high-frequency small bets. A ``wet'' board (e.g., J$\heartsuit$T$\heartsuit$8$\clubsuit$) distributes equity more evenly and demands larger, less frequent bets.

Our context engine classifies board texture along multiple dimensions: flush potential, straight connectivity, pair structure, and high-card composition. Special board types (e.g., monotone flops, four-to-a-flush turns) receive complete strategy overrides with pre-specified ATT/DEF budgets, because these rare textures require fundamentally different approaches. For LLMs, board texture classification eliminates a combinatorial reasoning burden. Models frequently err on this assessment (e.g., calling a two-tone connected board ``dry''), which cascades into incorrect strategy selection. By providing the texture label deterministically, we ensure the LLM starts from a correct premise.

\paragraph{Hand strength classification (P4 hand strength).}
Evaluating one's hand relative to the board and the likely opponent range is the most complex per-decision task in poker. The same hand (e.g., top pair with a king kicker) can be a strong value hand on a dry board in a single-raised pot, or a marginal bluff-catcher on a wet board in a 3-bet pot.

We define 23 hand-class categories covering two distinct subsystems. The \emph{made-hand system} (15 classes) assigns ATT/DEF budgets based on showdown value: strong hands receive high budgets for multi-street aggression, while marginal hands receive limited budgets that constrain them to checking or single-street defense. The \emph{drawing-hand system} (8 classes) operates differently: draws receive ATT budgets for semi-bluffing, but their defense is governed by \emph{pot-odds thresholds} specifying the maximum callable bet size per street, because draw equity depends on remaining cards rather than accumulated pressure. When a hand has both made-hand value and a draw (e.g., a pair plus a flush draw), the system applies a \emph{combo rule} that adds the draw's contribution to the made-hand baseline. Full classification tables and budget details are in Appendix~\ref{sec:budget-details}.

\paragraph{River blocker guidance (P5 Blocker).}
On the river, all draws have completed or missed and showdown value is fixed. The dominant strategic factor shifts to \emph{blocker effects}: holding a card that reduces the opponent's combinations of a strong hand makes bluffing more profitable, while \emph{not} blocking the opponent's bluff range makes calling more profitable. These interactions are well-documented in poker theory but difficult for LLMs to apply correctly, because evaluating blockers requires joint reasoning over one's own hand, the board, and the opponent's likely holdings.

P5 Blocker provides river-specific guidance on two complementary decisions: (1) \emph{when to bluff}: prioritizing hands with no showdown value that block the opponent's value range and maintain a consistent betting narrative; and (2) \emph{when to bluff-catch}: prioritizing hands with showdown value that do not block the opponent's bluffs, calibrated to minimum defense frequency. This module is activated only on the river because blocker reasoning is less relevant on earlier streets where draw equity and future card interaction dominate the decision.

\subsection{Attack/Defense Budget System}
\label{sec:budget-system}

The budget system translates context into action constraints, motivated by a fundamental poker principle: \emph{hands have finite strategic capacity}. A medium-strength hand can profitably absorb one bet but not three consecutive barrels. Expert human players possess clear intuitions about how many streets each hand can bet for value (attack) and how many streets it can call against aggression (defend). For instance, a top-pair hand can typically bet two streets and defend all three, while a middle pair can only defend one or two streets before folding becomes correct. These intuitions, accumulated through years of play and solver study, constitute a form of experiential knowledge that is difficult for LLMs to apply consistently. The ATT/DEF budget system encodes this expert knowledge as explicit numeric constraints: each hand class receives a budget specifying how many streets of aggression or defense it can sustain, and the budget depletes as the action line develops.

Each hand-in-context receives:
\begin{align}
B_{\mathrm{rem}}^{\mathrm{att}} &= B^{\mathrm{att}}(h,c) - \textstyle\sum_t w(a_t), \\
B_{\mathrm{rem}}^{\mathrm{def}} &= B^{\mathrm{def}}(h,c) - \textstyle\sum_t w(a_t),
\end{align}
where $B^{\mathrm{att}}(h,c)$ and $B^{\mathrm{def}}(h,c)$ are base budgets determined by hand class and context, and $w(a_t)$ is the pressure weight of each prior bet, proportional to bet size relative to the pot. The budget encodes three GTO insights without equilibrium computation: (i) geometric betting distributes pressure across streets, (ii) minimum defense frequency governs continuation decisions, and (iii) low-SPR situations simplify to commit-or-fold.

Base budgets are assigned by hand class, ranging from $\infty$ (nuts) down to 0 (trash). Actual budgets are reduced by context modifiers: wet/flush-possible boards, weaker kickers, higher pot types, and out-of-position play. Drawing hands use pot-odds thresholds for defense instead of DEF budgets. The complete budget tables are in Appendix~\ref{sec:budget-details}.

The budget system serves a dual purpose for LLMs. First, it provides a \emph{single scalar summary} of a complex multi-factor assessment. Second, it enforces \emph{multi-street coherence}: because the budget depletes with each bet, a hand that bets the flop has less budget remaining for the turn, naturally implementing the geometric pressure distribution that characterizes expert play. Without this mechanism, LLMs tend to make locally reasonable but globally incoherent decisions.

\paragraph{From budget to viable actions.}
The remaining budget, combined with the current action line, street, position, SPR, board texture, and hand properties, determines a set of viable actions presented to the LLM. This mapping accounts for position-specific raise thresholds, street-dependent sizing, low-SPR commitment/slow-play logic, draw-specific defense thresholds, and role-dependent defaults. The full viable action logic is detailed in Appendix~\ref{sec:budget-details}; here we emphasize that the system typically presents the LLM with a constrained choice among reasonable options rather than the full action space, and the LLM exercises judgment within these options. 

\subsection{Action Grounding}
\label{sec:grounding}

After context computation and skill retrieval, PokerSkill builds the final prompt containing: game state, selected skill entries (P1--P5 fragments relevant to this decision), computed context verdicts, viable actions with sizing ranges, and a structured output schema. The LLM returns structured JSON via forced tool-use, and a validator ensures the action is legal and the size falls within bounds. If the LLM output fails validation, the system defaults to the most conservative viable action; such fallbacks occur in fewer than 0.1\% of hands. The complete system is deterministic except for the LLM call itself, making it fully auditable and reproducible given the same LLM API access and game state inputs.

\section{Experiments}
\label{sec:experiments}

We evaluate PokerSkill against GTOWizard, a current state-of-the-art GTO benchmark for HUNL that provides AIVAT variance-reduced evaluation~\citep{burch2018aivat}. In its own published evaluation, GTOWizard beats the 2018 ACPC champion Slumbot \citep{jackson2013slumbot} by $194 \pm 41$ mbb/hand ~\citep{gtowizardbenchmark2026}.

\subsection{Setup}
\label{sec:eval-setup}

\paragraph{Models and protocol.}
We test three frontier LLMs with PokerSkill: GPT-5.5 with extended high reasoning (XHigh), Claude Opus 4.6 with maximum thinking, and Claude Opus 4.7 with maximum thinking. For the default-prompt baseline, the same models receive only game state and a generic poker instruction. We also evaluate a rule-based ablation that uses the PokerSkill skill library with deterministic action selection (always choosing the first viable action from the budget system) and no language model, isolating the contribution of the skill library from LLM reasoning. At each decision point, the context engine recomputes features, retrieves relevant P1--P5 fragments, queries the LLM, and validates the structured response. Conversation context is maintained within each hand. No agent queries a solver during play.

\paragraph{Sample sizes and settings.}
All PokerSkill and default prompt experiments are evaluated for at least 5{,}000 hands against GTOWizard with AIVAT variance reduction. 
Both PokerSkill agents use forced tool-use output (structured JSON) with temperature 1.0 (required by extended-thinking APIs). Each hand is dealt by the GTOWizard server. GPT-5.5 XHigh costs ${\sim}\$0.30$/hand; Claude Opus 4.6 and 4.7 cost ${\sim}\$0.07$/hand. All experiments were conducted between April and May 2026 using the then-current API versions of each model. Results may vary with future model updates.

\subsection{Main Results}
\label{sec:main-results}

\begin{table}[!ht]
\centering
\begin{tabular}{llr}
\toprule
\textbf{Agent} & \textbf{Method} & \textbf{mbb/hand} \\
\midrule
GPT-5.5 XHigh & PokerSkill & $-57 \pm 21$ \\
Claude Opus 4.6 & PokerSkill & $-80 \pm 29$ \\
Claude Opus 4.7 & PokerSkill & $-87 \pm 64$ \\
\midrule
Rule-based (no LLM) & PokerSkill Only & $-132 \pm 19$ \\
\midrule
GPT-5.5 XHigh & Default Prompt & $-132 \pm 25$ \\
GPT-5.3 XHigh & Default Prompt & $-160 \pm 30$ \\
Claude Opus 4.7 & Default Prompt & $-170 \pm 28$ \\
GPT-5.4 XHigh & Default Prompt & $-178 \pm 37$ \\
Slumbot (ACPC 2018) & Solver-based & $-194 \pm 41$ \\
Claude Opus 4.6 & Default Prompt & $-204 \pm 44$ \\
Claude Opus 4.5 & Default Prompt & $-223 \pm 51$ \\
\bottomrule
\end{tabular}
\caption{AIVAT performance against GTOWizard (mbb/hand). All PokerSkill experiments use $\geq$5{,}000 hands with AIVAT; default-prompt baselines are from the GTOWizard benchmark~\citep{gtowizardbenchmark2026} (GPT-5.3, GPT-5.4, Claude Opus 4.5, Claude Opus 4.6) or our reproduction using the same opponent (GPT-5.5, Claude Opus 4.7). ``Rule-based (no LLM)'' uses the PokerSkill skill library with deterministic action selection and no language model. $\pm$ denotes 1 SE.}
\label{tab:main-results}
\end{table}

\begin{figure}[t]
    \centering
    \includegraphics[width=0.72\textwidth]{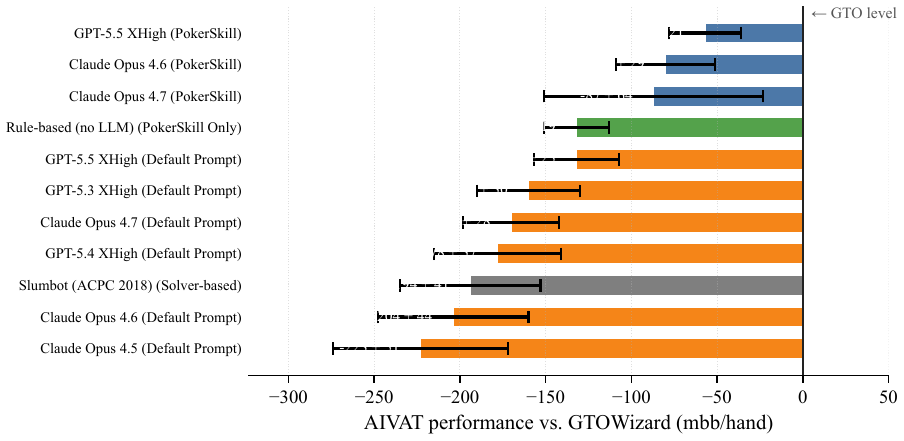}
    \caption{AIVAT loss rate against GTOWizard Benchmark. PokerSkill substantially closes the gap between raw LLM play and near-GTO performance.}
    \label{fig:gtowizard-results}
\end{figure}

Table~\ref{tab:main-results} reports AIVAT performance. GPT-5.5 XHigh improves from $-132$ to $-57$ mbb/hand with PokerSkill (57\% reduction). Claude Opus 4.6 improves from $-204$ to $-80$ mbb/hand (61\% reduction). Claude Opus 4.7 improves from $-170$ to $-87$ mbb/hand (49\% reduction). All three PokerSkill agents achieve lower loss rates than Slumbot ($-194 \pm 41$). We note that the Slumbot comparison is indirect. Nonetheless, the magnitude of the gap ($-57$ vs.\ $-194$) substantially exceeds measurement uncertainty.

Notably, the rule-based agent using only the PokerSkill skill library without any LLM achieves $-132 \pm 19$ mbb/hand. This is comparable to default-prompt LLMs ($-132$ to $-223$), demonstrating that the skill library alone captures substantial poker knowledge. However, the large gap between the rule-based agent ($-132$) and PokerSkill with GPT-5.5 ($-57$) confirms that LLM reasoning contributes meaningfully beyond the rule engine---particularly in ambiguous situations where the budget system permits multiple viable actions.

\subsection{Analysis}
\label{sec:analysis}

\paragraph{Model scaling and statistical significance.}
GPT-5.5 XHigh ($-57 \pm 21$) outperforms both Claude Opus 4.6 ($-80 \pm 29$) and Claude Opus 4.7 ($-87 \pm 64$) under the same framework. The difference between GPT-5.5 and Claude Opus 4.6 is not statistically significant at current sample sizes, while Claude Opus 4.7 performs comparably to Opus 4.6. We emphasize that the \emph{within-model} improvement from default prompting to PokerSkill is large in magnitude for all three models: GPT-5.5 improves by 75 mbb/hand, Claude Opus 4.6 by 124 mbb/hand, and Claude Opus 4.7 by 83 mbb/hand.

\paragraph{Default-prompt scaling paradox.} Under default prompting, performance does not scale monotonically with model capability: GPT-5.5 XHigh ($-132$) outperforms GPT-5.3 ($-160$), but GPT-5.4 ($-178$), Claude Opus 4.7 ($-170$), and older Claude models ($-204$ to $-223$) perform worse. This suggests that the decision-binding problem may worsen with increased reasoning depth if the model considers more factors without a mechanism to prioritize among them.


\section{Discussion and Conclusion}
\label{sec:discussion}

\paragraph{Why does structured scaffolding work?}
Claude Opus 4.6 with default prompting loses $-204$ mbb/hand; the same model with PokerSkill loses only $-80$ mbb/hand. The weights are identical---only the information presented at each decision changes. This is consistent with the hypothesis that a major bottleneck is the interface between stored knowledge and situated action, compounded by state-grounding and action-space errors that the deterministic engine resolves. The budget system produces viable actions but does not rank them; the LLM determines the final choice between them. The gap between GPT-5.5 XHigh ($-57$) and Claude Opus 4.6 ($-80$) under the identical framework suggests that model capability matters beyond the rule engine, though the relative contribution of the expert system versus LLM judgment is not isolated. PokerSkill should be understood as a case study in context-conditioned skill activation for LLM agents: the general pattern, including deterministic state abstraction, locally relevant expert retrieval, action-space bounding, and structured validation, applies broadly. We discuss implications in Appendix~\ref{sec:broader-impact}.

\paragraph{Conclusion.}
PokerSkill enables pure LLM agents to achieve competitive HUNL play without solver queries or game-tree traversal at inference time. GPT-5.5 XHigh achieves $-57 \pm 21$ mbb/hand against GTOWizard (57\% loss reduction); Claude Opus 4.6 improves by 59\% and Opus 4.7 by 49\%, demonstrating cross-family generality. By externalizing expert players' cognitive pipeline into a deterministic scaffold, PokerSkill addresses state-grounding, strategic binding, and action-validity failures simultaneously without modifying model weights. The skill library encodes expert knowledge informed by GTO theory; no solver is queried during play. We provide sufficient detail for independent replication and open-source code. Cross-model differences suggest the framework benefits from stronger base models, though scaling is not monotonic. Importantly, PokerSkill demonstrates a fundamentally different path from the dominant paradigm of training stronger poker-specific models or deploying real-time solvers: as foundation LLMs continue to improve, training-free poker agents may naturally achieve stronger performance without any additional engineering. Our results also suggest that rule-based AI need not be confined to the role of a weak baseline; when paired with a sufficiently capable reasoning engine, expert rules can serve as an effective interface for aligning general-purpose intelligence to specialized action spaces. Future work includes multiplayer settings and automated budget calibration.

\bibliographystyle{plainnat}
\bibliography{references}

\appendix
\newpage
\section{HUNL Rules and Terminology}
\label{sec:hunl-rules}

Heads-up No-Limit Texas Hold'em (HUNL) is a two-player poker game. Each player receives two private hole cards. Five public board cards are revealed over four betting streets: preflop, flop, turn, and river. The small blind posts half a big blind and acts first preflop; the big blind posts one big blind. After the flop, the button is in position (IP) and acts last, while the big blind is out of position (OOP) and acts first. Positions alternate every hand: the player who was the button becomes the big blind in the next hand, and vice versa.

Legal actions are \emph{fold}, \emph{check}, \emph{call}, \emph{bet}, \emph{raise}, and \emph{all-in}. Checking is legal only when facing no bet. Calling matches the current bet. Betting or raising requires a legal amount within the environment's minimum and maximum bounds. No-limit means the amount can vary up to the effective stack.

We report poker outcomes in big blinds (BB). BB/100 is big blinds won per 100 hands. GTOWizard reports AIVAT-adjusted milli-big-blinds per hand (mbb/hand), where $1$ bb/100 equals $10$ mbb/hand. AIVAT is a variance-reduction estimator for poker evaluation \citep{burch2018aivat}; Slumbot does not provide it, so Slumbot results are raw chip outcomes with wider confidence intervals. All confidence intervals (CI) reported in this paper are 95\% confidence intervals computed as $\bar{x} \pm 1.96 \cdot \mathrm{SE}$, where $\mathrm{SE} = s / \sqrt{n}$ is the standard error of the AIVAT-adjusted per-hand outcomes. GTO denotes game-theoretic optimal or approximate equilibrium play. A range is a distribution over possible private hands. Board texture describes how coordinated the public cards are. A blocker is a private card that reduces the opponent's combinations of a strong hand. A bluff-catcher is a hand that usually loses to value bets but can beat bluffs. SPR is stack-to-pot ratio, the effective remaining stack divided by the pot.

\section{A Brief History of HUNL and Poker AI}
\label{sec:hunl-history}

HUNL has served as a primary benchmark for imperfect-information game AI for over two decades. The evolution of HUNL agents illustrates a progression from interpretable but weak heuristics to powerful but opaque computational systems. PokerSkill represents a distinct point in this landscape: a training-free, solver-free agent that leverages rule-based knowledge through modern language models.

\paragraph{Era 1: Rule-based and expert systems (1990s--2005).}
The earliest poker AI programs used hand-crafted rules for betting decisions. The University of Alberta's poker research group developed Loki~\citep{billings2002challenge}, which combined hand strength estimation, opponent modeling via neural networks, and rule-based betting logic. These systems were interpretable and required no large-scale computation, but their strategic depth was limited by the expressiveness of manually authored rules. They could not represent the complex conditional strategies needed for equilibrium play in no-limit variants.

\paragraph{Era 2: Abstraction and CFR (2005--2015).}
The introduction of counterfactual regret minimization (CFR)~\citep{zinkevich2007regret} transformed poker AI. By iteratively minimizing regret over the game tree, CFR converges to Nash equilibrium in two-player zero-sum games. However, full HUNL has $\sim$10$^{164}$ information sets~\citep{johanson2013measuring}, far exceeding direct computation. Researchers developed action abstraction~\citep{ganzfried2014potential} and card abstraction to reduce the game to tractable size, then solved the abstracted game with CFR variants~\citep{lanctot2009monte,brown2019solving}. Slumbot~\citep{jackson2013slumbot} exemplifies this era: a strong HUNL agent built on large-scale CFR with abstraction, which won the 2018 Annual Computer Poker Competition.

\paragraph{Era 3: Depth-limited solving and real-time computation (2016--present).}
DeepStack~\citep{moravcik2017deepstack} and Libratus~\citep{brown2018superhuman} achieved superhuman HUNL play by combining CFR with depth-limited solving~\citep{brown2018depth} and subgame re-solving~\citep{brown2017safe}. Rather than solving the entire game offline, these systems solve subgames in real time, using neural networks (DeepStack) or precomputed blueprints (Libratus) to estimate values beyond the solving depth. This approach requires millions of CPU core-hours for blueprint computation and significant real-time compute per decision. GTOWizard~\citep{gtowizardbenchmark2026} represents the current state-of-the-art benchmark in this paradigm, beating Slumbot by $194 \pm 41$ mbb/hand. EVPA~\citep{li2025evpa}, RL-CFR~\citep{li2024rlcfr}, and TurboReBeL~\citep{li2026turborebel} further improve abstraction and solving efficiency, enabling stronger equilibrium approximation with less computation.

\paragraph{Era 4: LLM-based agents (2023--present).}
The emergence of frontier language models opened a new direction: using LLMs as poker decision-makers without game-theoretic training. Early evaluations~\citep{gupta23,zhuang2025pokerbench} found that even GPT-4 deviates substantially from GTO play. ToolPoker~\citep{lin2026how} queries external solvers to improve LLM decisions but still depends on solver access. PokerSkill demonstrates that a pure LLM agent, equipped with structured rule-based skills but no solver or training, can achieve performance within the range of solver-based systems against a state-of-the-art benchmark.

\paragraph{Comparative positioning.}
Table~\ref{tab:poker-ai-comparison} summarizes the key characteristics of representative HUNL agents across these eras.

\begin{table}[h]
\centering
\caption{Comparison of HUNL poker AI approaches. ``Training'' refers to any offline computation (CFR iterations, RL, fine-tuning). ``Interpretable'' means the decision logic can be inspected by humans.}
\label{tab:poker-ai-comparison}
\begin{tabular}{lcccc}
\toprule
\textbf{System} & \textbf{Training} & \textbf{Interpretable} & \textbf{Scales with LLM} & \textbf{vs.\ GTOWizard} \\
\midrule
Loki (rule-based) & None & Yes & N/A & $\gg$-200 mbb/h \\
Slumbot (CFR) & Millions CPU-hrs & No & No & $-194 \pm 41$ mbb/h \\
DeepStack & Millions CPU-hrs & No & No & Near-human \\
Libratus & Millions CPU-hrs & No & No & Superhuman \\
GTOWizard & Large-scale & No & No & (reference) \\
GPT-5.5 (zero-shot) & None & Partial & Yes & $-132 \pm 25$ mbb/h \\
\textbf{PokerSkill} & \textbf{None} & \textbf{Yes} & \textbf{Yes} & $\mathbf{-61 \pm 66}$ \textbf{mbb/h} \\
\bottomrule
\end{tabular}%
\end{table}

PokerSkill's unique position is that it requires \emph{zero training and zero solver access}, yet achieves results within the confidence interval of much more computationally expensive systems. Moreover, because the decision quality depends on the underlying LLM's reasoning capability, PokerSkill's performance is expected to improve as foundation models advance---a property not shared by fixed solver-based systems.

\section{Prompt Layers}
\label{sec:prompt-details}

PokerSkill uses five prompt layers. P1 contains the game rules, output schema, and legal-action protocol. P2 contains preflop range guidance. P3 contains general postflop principles such as position, pot control, value/bluff separation, and bet sizing. P4 contains context-specific strategy keyed by board texture, action line, hand class, and role. P5 Blocker contains river bluff and bluff-catch guidance. The main paper describes the selection mechanism; this appendix records the implementation-facing taxonomy.

\subsection{P1: System and Output Protocol}

P1 is always included. It specifies that the model is playing HUNL, must choose only from legal actions, must respect the listed viable options, and must return a structured JSON action. The essential output schema is \texttt{\{action, amount\}}: \texttt{action} is one of fold, check, call, bet, raise, or all-in, and \texttt{amount} is a number or null. P1 also states that strategic reasoning should follow the computed ATTACK and DEFENSE budgets rather than inventing unconstrained bet sizes.

A minimal P1-style instruction is:
\begin{quote}\small
You are a heads-up no-limit Texas hold'em agent. Use the situation analysis and selected poker skills below. Choose exactly one action from the legal and viable options. If betting or raising, output a legal amount. Return JSON only.
\end{quote}

\subsection{P2: Preflop Ranges}

P2 is selected only before the flop. The preflop scenario detector maps the action history to cases such as unopened button, big blind versus open, 3-bet pot, 4-bet pot, squeeze-like line, short-stack all-in threshold, or limp pot. The selected range text gives the model compact advice for opening, calling, 3-betting, 4-betting, and folding. Only the relevant range table is included; the model is not shown all preflop charts at every decision.

\subsection{P3--P5 Blocker: Postflop Skills}

P3 contains stable postflop principles: separate value from bluffs, avoid betting marginal showdown hands without a reason, prefer position when realizing equity, use smaller bets on static boards, use protection bets on dynamic boards, and avoid illegal or meaningless bet sizes. P4 is retrieved by matching board texture, hero hand class, action-line scenario, position, and role. It gives local attack guidance, defense guidance, and warnings such as one-card flush danger or dominated kicker risk. P5 Blocker is included on the river and emphasizes that draws no longer have equity, blockers become more important, and marginal showdown hands should not be turned into value bets.

\subsection{Prompt Template}
\label{sec:prompt-template}

A simplified postflop prompt has this form:
\begin{quote}\small
\textbf{Situation.} Street, pot, stacks, legal actions, hero cards, board, action history.\\
\textbf{Computed context.} Pot type, position, initiative, SPR, board texture, hand class, draw class, action-line scenario, weighted pressure.\\
\textbf{Budget verdict.} Attack budget, defense budget, remaining budget, and reason.\\
\textbf{Selected skills.} P3 general principles; P4 entries for this board/hand/line; P5 river guidance if river.\\
\textbf{Viable options.} A short enumerated list of allowed strategic choices.\\
\textbf{Output.} Return JSON with one action and amount.
\end{quote}


\section{Context Feature Taxonomy}
\label{sec:context-taxonomy}

The context engine computes features in four groups. \textbf{Board features} include dry/wet texture, paired boards, trips boards, monotone boards, flush-completing boards, straight-completing boards, and one-card straight or flush threats. \textbf{Hand features} include made-hand class, draw class, kicker class, and high-card showdown labels. \textbf{Line features} include pot type, initiative, position, street, prior bet count, weighted bet pressure, and scenario ID. \textbf{Stack features} include SPR and all-in thresholds.

\subsection{Board Analysis and Special Boards}

The board analyzer classifies public cards by texture and special structure. Texture labels include \texttt{dry}, \texttt{slightly\_wet}, \texttt{wet}, and \texttt{very\_wet}. Suit labels include two-tone, monotone, flush-possible, and one-card-flush danger. Rank labels include paired, trips, double-paired, quads, straight-completing, and one-card straight threats. These labels directly modify budgets and determine which P4 warnings are retrieved.

Special boards are handled separately because normal hand-ordering rules can be misleading. On paired or trips boards, a nominally strong hand can be dominated by full houses. On double-paired boards, kicker rank can decide whether a high-card hand has showdown value. On quads or board-full-house textures, private-card kickers or exact rank interaction can matter more than ordinary pair categories. On one-card flush or one-card straight boards, a single private card can complete a very strong hand, so medium showdown and high-card defense can be reduced to zero.

Examples:
\begin{itemize}
    \item $K\spadesuit 7\diamondsuit 2\clubsuit$ is dry and high-card dominated.
    \item $9\clubsuit 8\clubsuit 6\diamondsuit$ is wet because it contains both flush and straight pressure.
    \item $4\clubsuit 9\clubsuit 4\spadesuit$ is paired and two-tone; high-card showdown can gain limited bluff-catching value, but weak kickers remain fragile.
    \item A four-flush board activates one-card flush danger, sharply reducing high-card and medium-showdown defense.
\end{itemize}

\subsection{Hand Analysis}

The hand analyzer returns an absolute strength class and a draw class. Absolute strength includes \texttt{nuts}, \texttt{flush}, \texttt{straight}, \texttt{set}, \texttt{trip}, \texttt{two\_pair}, \texttt{overpair}, \texttt{top\_pair}, \texttt{second\_pair}, \texttt{third\_pair},  \texttt{lower\_pair}, \texttt{nuts\_high},  \texttt{second\_high} and no-showdown trash. Draw strength includes \texttt{strong\_draws}, \texttt{medium\_strong\_draws}, \texttt{medium\_draws}, \texttt{medium\_weak\_draws}, \texttt{weak\_draws}, \texttt{strong\_overcard\_draws}, \texttt{medium\_overcard\_draws}, \texttt{weak\_overcard\_draws}.





\section{Attack/Defense Budget Details}
\label{sec:budget-details}

The budget system maps previous betting pressure to a scalar cost. Each hand class receives base attack and defense budgets, then pot type, board texture, position, SPR, and special-board modifiers adjust the budget. The action-line module compares the remaining budget to thresholds and emits a verdict plus viable actions. Close thresholds use fuzzy zones so the prompt can preserve both options rather than making discontinuous recommendations.

\subsection{Attack and Defense Budget}

Attack budget controls proactive betting and raising. Strong value hands receive high attack budget. Vulnerable but strong hands can retain attack budget on wet boards because protection has value. Marginal showdown and high-card showdown hands receive little or no attack budget because they should not bet for value. Draws can receive attack budget when fold equity and equity realization make bluffing reasonable, especially with strong draws or combo draws.

A reproduction should implement attack logic with three steps: assign a base attack budget from hand class, adjust by board danger, position, initiative, and SPR, and subtract weighted betting pressure already invested in the line. If remaining attack budget is clearly positive, betting or raising can appear in viable options. If it is near zero, only small bets or checks may remain. If it is negative, betting is removed unless an override applies for very strong hands or low-SPR commitment nodes.


Defense budget controls calls and defensive raises when facing bets. It starts from hand class and is modified by pot type, kicker, board danger, and position. Medium showdown hands defend more in limped and single-raised pots than in 3-bet and 4-bet pots. One-card flush or one-card straight boards can reduce defense to zero for high-card showdown hands. Paired boards can slightly increase showdown value, but this bonus does not turn dominated high-card hands into automatic calls.


\subsection{Budget Tables and Prompt Taxonomy}

The prompt exposes the following implementation-facing tables. Values are weighted-bet budgets before line-pressure subtraction. ``Unlimited'' is represented by a sentinel value in the implementation; ranges indicate kicker, rank, or texture-dependent variants.

\paragraph{Board texture and special-board retrieval.}
\begin{center}\small
\setlength{\tabcolsep}{4pt}
\begin{tabular}{p{0.22\columnwidth}p{0.66\columnwidth}}
\toprule
Feature & Prompt effect \\
\midrule
\texttt{dry}/\texttt{very\_dry} & Small high-frequency c-bets, stable made-hand values, thinner value possible. \\
\texttt{wet}/\texttt{very\_wet} & Larger sizes, lower c-bet frequency, protection and draw interaction emphasized. \\
Two-tone / 3-flush / 4-flush & Activates flush-draw, flush-complete, or one-card-flush warnings; non-flush made hands are downgraded. \\
Straight possible / one-card straight & Downgrades one-pair and two-pair hands; one-card straights distinguish top-end from low-end. \\
Paired board & Trips/full-house risk; paired-board penalties for pairs, limited bonus for high-card showdown. \\
Trips board & Quads/full-house/kicker override table replaces normal pair logic. \\
Double-paired / board full house & Private-card rank interaction decides whether hero has nuts, lower full house, shared board hand, or kicker-only showdown. \\
Board straight / board flush & Shared-board logic; private blocker or suited-card rank determines whether hero improves beyond the board. \\
\bottomrule
\end{tabular}
\end{center}

\paragraph{Complete hand-class ATT/DEF prompt content.}
The following is the exact prompt content provided to the LLM for each hand class. Each entry specifies the base ATT/DEF budgets and all context-dependent modifiers.

\medskip\noindent\textbf{Hand class: nuts.}\\
\begin{lstlisting}
Full house+ / Nut flush / Nut straight (no-flush no-pair board) / Set (no-flush no-pair no-straight board):
  ATTACK: unlimited. DEFENSE: unlimited (always call/raise).
  Build pot aggressively. Consider overbet on nut advantage boards. Low SPR -> can slow-play one street to trap. RIVER: if you are last to act and have a raise option, ALWAYS raise -- calling with the nuts loses EV.
\end{lstlisting}

\medskip\noindent\textbf{Hand class: flush.}\\
\begin{lstlisting}
Flush:
  [3-FLUSH BOARD (3 same suit)]:
    Nut flush: ATT/DEF unlimited.
    Big flush (high card >9): ATT 5 / DEF 6.
    Small flush (high card <=9): ATT 4 / DEF 5.
    [PAIRED BOARD]: Nut flush ATT 5 DEF 6. Big flush ATT 4.3 DEF 5.3. Small flush ATT 3.5 DEF 4.5.
  [ONE-CARD FLUSH BOARD (4+ same suit)]:
    Nut: ATT/DEF unlimited. 2nd: ATT 4 DEF 5. 3rd: ATT 3 DEF 4.
    4th: ATT 2.4 DEF 3.5. 5th: ATT 2 DEF 3. 6-7th: ATT 1.5 DEF 2.5. 8-9th: ATT 1 DEF 2.
    [PAIRED BOARD]: Nut flush ATT 4.5 DEF 5.5; others ATT/DEF -0.5.
\end{lstlisting}

\medskip\noindent\textbf{Hand class: straight.}\\
\begin{lstlisting}
Straight:
  TWO-CARD STRAIGHT (both hole cards contribute):
    No flush: ATT 5.5 DEF 6.5. Flush possible (3-flush): ATT 3.5 DEF 4.5.
    4+ flush board: ATT 0 DEF 1.
    [PAIRED BOARD]: Nut straight + no flush -> ATT 5 DEF 6. Otherwise ATT/DEF -0.4.
  ONE-CARD STRAIGHT (1 hole card contributes):
    TOP-END (nut straight):
      No flush: NUTS. 3-flush: ATT 2.5 DEF 3.5. 4+flush: ATT 0 DEF 1.
    LOW-END (higher straight possible):
      No flush: ATT 2.5 DEF 3.5. 3-flush: ATT 1.5 DEF 2.5. 4+flush: ATT 0 DEF 0.5.
    [PAIRED BOARD]: ATT/DEF -0.6.
\end{lstlisting}

\medskip\noindent\textbf{Hand class: set.}\\
\begin{lstlisting}
Set (pocket pair hits board):
  No flush, no straight possible: NUTS (ATT/DEF unlimited).
  No flush, has 2-card straight: 1 possibility ATT 5.5 DEF 6.5; 2 -> ATT 4.5 DEF 5.5; 3+ -> ATT 4 DEF 5.
  3-flush board (no OCS): ATT 3.8 DEF 4.8, each additional straight possibility -0.3.
  One-card straight (OCS) 1 type: ATT 2.5 DEF 3.7. OCS 2+ types: ATT 1.5 DEF 2.7.
    OCS + 3-flush: ATT/DEF -0.8 from OCS base. OCS + 4+flush: ATT 0 DEF 0.5.
  4+ flush board (no OCS): ATT 0 DEF 1.
\end{lstlisting}

\medskip\noindent\textbf{Hand class: trips.}\\
\begin{lstlisting}
Trips (board pair + one hole card):
  Dry board (no flush/OCS): ATT 4(2-kicker)-4.5(A-kicker) DEF 5.5, each 2-card str -0.3.
  3-flush board (no OCS): ATT 3-3.5 DEF 4.5, each 2-card str -0.3.
  OCS 1 type: ATT 2.1-2.4 DEF 3.5. OCS 2+ types: ATT 1.1-1.4 DEF 2.5.
    OCS + 3-flush: ATT/DEF -0.7 from OCS base. OCS + 4+flush: ATT 0 DEF 0.5.
  4+ flush board (no OCS): ATT 0 DEF 1.
  Kicker scales ATT proportionally: lowest(2)->base, highest(A)->max.
\end{lstlisting}

\medskip\noindent\textbf{Hand class: two\_pair.}\\
\begin{lstlisting}
Two pair (NO board pair):
  ATT/DEF by board texture (condition-matrix, highest-priority match):
  [4+ FLUSH (OCF) + ONE-CARD STRAIGHT]: ATT 0 / DEF 0.5.
  [4+ FLUSH (OCF), no OCS]: ATT 0 / DEF 1.
  [ONE-CARD STRAIGHT + 3-FLUSH]: 1 type OCS -> ATT 1.7 / DEF 2.8; 2+ types -> ATT 0.7 / DEF 1.9.
  [ONE-CARD STRAIGHT, no flush]: 1 type OCS -> ATT 2.2 / DEF 3.3; 2+ types -> ATT 1.2 / DEF 2.4.
  [3-FLUSH, no OCS]: rank-based ATT 2.7(r10)-3.6(r1) / DEF 3.7(r10)-4.7(r1); each 2-card straight -0.3.
  [DRY BOARD]: Rank 1 (top): ATT 5 / DEF 6.5. R2: 4.7/6. R3: 4.5/5.7. R4: 4.3/5.5.
    R5: 4.1/5.3. R6: 3.9/5.1. R7: 3.8/5. R8: 3.7/4.9. R9: 3.6/4.8. R10: 3.5/4.7.
    Each 2-card straight possibility -0.3 ATT/DEF.
  Top two pair on safe boards = go for stacks over all streets. Bottom two pair = still aggressive, but cautious on scary turn/river cards.
  NOTE: If board has a pair, your 'two pair' includes the shared board pair -- real strength ~ top pair/middle pair (see your classification above).
\end{lstlisting}

\medskip\noindent\textbf{Hand class: overpair.}\\
\begin{lstlisting}
Overpair (pocket pair above all board cards):
  ATTACK / DEFENSE reference (by pot type and pair rank):
    SRP/Limp -- AA: ATT 3.5 / DEF 4.5. KK: ATT 3.4 / DEF 4.4. QQ: ATT 3.3 / DEF 4.3. JJ: ATT 3.2 / DEF 4.2. Others: ATT 3.1 / DEF 4.1.
    3BP+     -- AA: ATT 3.4 / DEF 4.5. KK: ATT 3.2 / DEF 4.3. QQ: ATT 3.0 / DEF 4.1. JJ: ATT 2.8 / DEF 3.8. TT: ATT 2.6 / DEF 3.7. Others: ATT 2.5 / DEF 3.5.
    4BP      -- AA: ATT 3.4 / DEF 4.5. KK: ATT 3.1 / DEF 4.2. QQ: ATT 2.7 / DEF 3.7. JJ: ATT 2.4 / DEF 3.4. Others: ATT 2.1 / DEF 3.1.
  [SAFE BOARD -- no flush possible (rainbow/two-tone), no 1-card straight, no straight possible, not paired]:
    SRP: very strong -- triple barrel for value. In 3BP: triple barrel safe boards. In 4BP: standard value, get stacks in.
  [PAIRED BOARD (board has a pair)]: Overpair on paired board: ATT/DEF -0.5. Still STRONG -- 2-3 streets of value. Flop paired board: c-bet and continue normally. Turn pairing: CHECK this specific street (opponent gains trips combos), but still bet the other streets. Overpair beats all one-pair hands and most bluffs -- do not play passively just because board is paired.
  [ONE-CARD FLUSH (board 4+ same suit, hero has NO flush)]: ATTACK 0, DEFENSE capped at 0.6. Near-worthless -- check/fold.
  [ONE-CARD STRAIGHT -- OPEN-ENDED]: drop 2.5 levels -> WEAK SHOWDOWN. Bet 1 street small at most, fold to heavy aggression.
  [ONE-CARD STRAIGHT -- GUTSHOT]: drop 1.5 levels -> THIN VALUE ONLY. Bet 1 street small, fold to heavy aggression.
  [FLUSH POSSIBLE BOARD (3+ same suit)]: drop 1.1(flop) / 0.9(turn) / 0.7(river) levels.
  [STRAIGHT POSSIBLE BOARD -- multiple combos]: drop 0.6(flop) / 0.5(turn) / 0.4(river) levels.
  [STRAIGHT POSSIBLE BOARD -- only 1 combo]: drop 0.4(flop) / 0.3(turn) / 0.2(river) levels.
  NOTE: board texture penalties STACK, EXCEPT: [FLUSH POSSIBLE] and [ONE-CARD FLUSH] do NOT stack (use more severe); [STRAIGHT POSSIBLE] and [ONE-CARD STRAIGHT] do NOT stack (use more severe).
  Facing check-raise -> re-evaluate on all board types.
\end{lstlisting}

\medskip\noindent\textbf{Hand class: top\_pair.}\\
\begin{lstlisting}
Top pair:
  ATTACK / DEFENSE reference (by pot type and kicker):
    SRP  -- TPTK: ATT 3 / DEF 4. TPSK: ATT 2.8 / DEF 3.8. 3rd kicker: ATT 2.6 / DEF 3.6. 4th: ATT 2.4 / DEF 3.4. 5th: ATT 2.2 / DEF 3.2. Other: ATT 2.1 / DEF 3.1.
    3BP  -- TPTK: ATT 2.9 / DEF 3.9. TPSK: ATT 2.6 / DEF 3.6. 3rd kicker: ATT 2.2 / DEF 3.2. Other: ATT 1.9 / DEF 2.9.
    4BP+ -- TPTK: ATT 2.6 / DEF 3.6. TPSK: ATT 2.2 / DEF 3.2. 3rd kicker: ATT 1.8 / DEF 2.8. Other: ATT 1.6 / DEF 2.6.
  [SAFE BOARD -- no flush possible (rainbow/two-tone), no 1-card straight, no straight possible, not paired]:
    SRP TPGK / 3BP TPTK-TPSK: triple barrel for value on safe boards. Defend normal-size triple barrel.
    Other kickers: bet 2 streets, check 1 street for pot control.
  [PAIRED BOARD]: ATT/DEF -0.5. Opponent has more trips/full-house combos. Turn pairing: CHECK that specific street (opponent gains trips combos), bet the other streets.
  [ONE-CARD FLUSH (4+ same suit, hero has NO flush)]: drop 3.5 levels -> WEAK SHOWDOWN. Check, maybe call 1 small bet.
  [ONE-CARD STRAIGHT -- OPEN-ENDED]: drop 2.5 levels -> nearly WEAK SHOWDOWN. Check, maybe call 1 small bet.
  [ONE-CARD STRAIGHT -- GUTSHOT]: drop 1.5 levels -> THIN VALUE. Bet 1 street small at most.
  [FLUSH POSSIBLE BOARD (3+ same suit)]: drop 1.1(flop) / 0.9(turn) / 0.7(river) levels.
  [STRAIGHT POSSIBLE BOARD -- multiple combos]: drop 0.6(flop) / 0.5(turn) / 0.4(river) levels.
  [STRAIGHT POSSIBLE BOARD -- only 1 combo]: drop 0.4(flop) / 0.3(turn) / 0.2(river) levels.
  NOTE: board texture penalties STACK, EXCEPT: [FLUSH POSSIBLE] and [ONE-CARD FLUSH] do NOT stack (use more severe); [STRAIGHT POSSIBLE] and [ONE-CARD STRAIGHT] do NOT stack (use more severe).
  Facing check-raise -> re-evaluate on all board types.
\end{lstlisting}

\medskip\noindent\textbf{Hand class: second\_pair.}\\
\begin{lstlisting}
Second pair (1 board overcard above hero's pair):
  ATTACK / DEFENSE reference (by pot type and kicker):
    SRP  -- Pocket pair/top kicker: ATT 1.8 / DEF 2.8. 2nd kicker: ATT 1.7 / DEF 2.7. 3rd kicker: ATT 1.6 / DEF 2.6. Other: ATT 1.5 / DEF 2.5.
    3BP  -- Pocket pair: ATT 1.3 / DEF 2.5. Top kicker: ATT 1.5 / DEF 2.3. 2nd kicker: ATT 1.3 / DEF 2.1. Other: ATT 1.2 / DEF 2.0.
    4BP+ -- Pocket pair: ATT 0 / DEF 2. Top kicker: ATT 1 / DEF 1.8. Other: ATT 0.9 / DEF 1.6.
  This hand has SOLID SHOWDOWN VALUE -- play for pot control.
  [PAIRED BOARD]: ATT/DEF -0.4.
  As aggressor: c-bet flop, bet 1 more street on safe boards. Max 2 streets of betting.
  River with second pair when double barreled before -> CHECK BACK.
  As defender SRP: check-call 2 streets. Facing <=40% pot bet -> call.
  In 3BP: check-call 1-2 streets, fold to continued heavy aggression.
  In 4BP: call once, fold to huge bets or continued aggression.
  [ONE-CARD FLUSH (4+ same suit, hero has NO flush)]: drop 3.5 levels -> Nearly TRASH.
  [ONE-CARD STRAIGHT -- OPEN-ENDED]: drop 2.5 levels.
  [FLUSH POSSIBLE BOARD (3+ same suit)]: drop 1.1(flop) / 0.9(turn) / 0.7(river) levels.
  [STRAIGHT POSSIBLE BOARD -- multiple combos]: drop 0.6(flop) / 0.5(turn) / 0.4(river) levels.
  [STRAIGHT POSSIBLE BOARD -- only 1 combo]: drop 0.4(flop) / 0.3(turn) / 0.2(river) levels.
  NOTE: board texture penalties STACK, EXCEPT: [FLUSH POSSIBLE] and [ONE-CARD FLUSH] do NOT stack (use more severe); [STRAIGHT POSSIBLE] and [ONE-CARD STRAIGHT] do NOT stack (use more severe).
  Facing check-raise -> re-evaluate on all board types.
\end{lstlisting}

\medskip\noindent\textbf{Hand class: third\_pair.}\\
\begin{lstlisting}
Third pair (2 board overcards above hero's pair):
  ATTACK / DEFENSE reference (by pot type and kicker):
    SRP  -- Top kicker/pocket pair: ATT 1.2 / DEF 2.2. Other: ATT 1.0 / DEF 2.0.
    3BP  -- DEFENSE 1.5.  4BP+ -- DEFENSE 1.2.
    3BP+ -- Board-hit pair: ATTACK 1.5 (stab <=30% pot when opp shows weakness, check river).            Pocket pair: ATTACK 0 DEFEND 0.6(3BP)/0.3(4BP+).
  3BP/4BP overcapped pocket pair (not hitting board, only 2 outs):
  [PAIRED BOARD]: ATT/DEF -0.3.
  This hand has MARGINAL SHOWDOWN VALUE.
  SRP: c-bet flop small (range bet ~33%), then CHECK remaining streets.
  3BP/4BP board-hit: can stab 1 street when opponent checks (showing weakness), then check river. Pocket pair: pure showdown.
  As defender SRP: facing c-bet -> call. Check-call 1-2 streets total. Fold to triple barrel.
  In 3BP/4BP: call once, fold to continued aggression.
  [ONE-CARD FLUSH (4+ same suit)]: drop 3.5 levels -> TRASH.
  [ONE-CARD STRAIGHT -- OPEN-ENDED]: drop 2.5 levels.
  [FLUSH POSSIBLE BOARD (3+ same suit)]: drop 1.1(flop) / 0.9(turn) / 0.7(river) levels.
  [STRAIGHT POSSIBLE BOARD -- multiple combos]: drop 0.6(flop) / 0.5(turn) / 0.4(river) levels.
  [STRAIGHT POSSIBLE BOARD -- only 1 combo]: drop 0.4(flop) / 0.3(turn) / 0.2(river) levels.
  NOTE: board texture penalties STACK, EXCEPT: [FLUSH POSSIBLE] and [ONE-CARD FLUSH] do NOT stack (use more severe); [STRAIGHT POSSIBLE] and [ONE-CARD STRAIGHT] do NOT stack (use more severe).
\end{lstlisting}

\medskip\noindent\textbf{Hand class: fourth\_fifth\_pair.}\\
\begin{lstlisting}
4th/5th pair (3-4 board overcards above hero's pair):
  ATTACK / DEFENSE reference (by pot type):
    SRP/Limp -- 4th pair: ATTACK 0.8 / DEFENSE 1.8. 5th pair: ATTACK 0.5 / DEFENSE 1.5.
    3BP  -- DEFENSE 1.0.  4BP+ -- DEFENSE 0.7.
    3BP+ -- Board-hit pair: ATTACK 1.5 (stab <=30% pot when opp shows weakness, check river).            Pocket pair: ATTACK 0 DEFENSE 0.3(3BP)/0(4BP+).
  3BP/4BP overcapped pocket pair (not hitting board, only 2 outs):
  [PAIRED BOARD]: ATT/DEF -0.3.
  This hand has MINIMAL SHOWDOWN VALUE -- barely above trash.
  3BP/4BP board-hit: can stab 1 street when opponent checks, then check river.
  NEVER bet >=90% for value.
  As defender SRP: facing c-bet -> can call once. Fold to double barrel. Fold to triple barrel.
  In 3BP/4BP: fold to first significant bet (>=70% pot).
  [ONE-CARD FLUSH (4+ same suit)]: drop 3.5 levels -> TRASH.
  [ONE-CARD STRAIGHT -- OPEN-ENDED]: drop 2.5 levels -> TRASH.
  [FLUSH POSSIBLE BOARD (3+ same suit)]: drop 1.1(flop) / 0.9(turn) / 0.7(river) levels.
  [STRAIGHT POSSIBLE BOARD -- multiple combos]: drop 0.6(flop) / 0.5(turn) / 0.4(river) levels.
  [STRAIGHT POSSIBLE BOARD -- only 1 combo]: drop 0.4(flop) / 0.3(turn) / 0.2(river) levels.
  NOTE: board texture penalties STACK, EXCEPT: [FLUSH POSSIBLE] and [ONE-CARD FLUSH] do NOT stack (use more severe); [STRAIGHT POSSIBLE] and [ONE-CARD STRAIGHT] do NOT stack (use more severe).
\end{lstlisting}

\medskip\noindent\textbf{Hand class: nuts\_high.}\\
\begin{lstlisting}
Nut high (A-high -- highest rank not on board):
  ATTACK 0. DEFENSE varies by pot type and kicker position (higher kicker -> more defense):
    Limp Pot: 0.8-1.2. SRP: 0.6-1.0. 3BP: 0.4-0.7. 4BP+: 0.1-0.4.
  Best non-pair showdown -- beats ALL bluffs and lower high cards.
  Do NOT bet for value. Check back or call small bets.
  PAIRED BOARD: DEFENSE boosted x1.35 (paired board = fewer pair combos for villain).
  [ONE-CARD FLUSH / ONE-CARD STRAIGHT]: DEFENSE -> 0 (too many hands beat you).
  [FLUSH POSSIBLE (3+ suit)]: penalty -1.0(flop) / -0.7(turn) / -0.4(river).
  [STRAIGHT POSSIBLE -- multi combos]: penalty -0.5(flop) / -0.4(turn) / -0.2(river).
  [STRAIGHT POSSIBLE -- 1 combo]: penalty -0.3(flop) / -0.2(turn) / -0.1(river).
\end{lstlisting}

\medskip\noindent\textbf{Hand class: second\_high.}\\
\begin{lstlisting}
Second high (K/Q-high -- 2nd-highest rank not on board, kicker T+ on unpaired board):
  ATTACK 0. DEFENSE varies by pot type and kicker position:
    Limp Pot: 0.4-0.7. SRP: 0.3-0.5. 3BP: 0.1-0.3. 4BP+: 0.
  Marginal showdown -- only beats bluffs. Do NOT bet.
  3BP+: fold to most bets.
  PAIRED BOARD: DEFENSE boosted x1.35.
  [ONE-CARD FLUSH / ONE-CARD STRAIGHT]: DEFENSE -> 0.
  [FLUSH POSSIBLE (3+ suit)]: penalty -1.0(flop) / -0.7(turn) / -0.4(river).
  [STRAIGHT POSSIBLE -- multi combos]: penalty -0.5(flop) / -0.4(turn) / -0.2(river).
  [STRAIGHT POSSIBLE -- 1 combo]: penalty -0.3(flop) / -0.2(turn) / -0.1(river).
\end{lstlisting}

\medskip\noindent\textbf{Hand class: weak\_showdown.}\\
\begin{lstlisting}
Weak showdown (low-kicker high card or very marginal unpaired hand):
  ATTACK / DEFENSE reference (by pot type):
    Limp Pot/SRP  -- ATTACK 0 / DEFENSE 0.8.
    3BP  -- ATTACK 0 / DEFENSE 0.4.
    4BP+  -- ATTACK 0 / DEFENSE 0.2
  Minimal showdown value. Do NOT bet for value.
  Limp Pot/SRP IP: check back for pot control, can call small bets (<=33% pot) on dry board.
  Limp Pot/SRP OOP: check-call at most once on dry board with small bet, fold to large or double barrel.
  In 3BP+: fold to >=70% bet (EXCEPTION: paired/double-paired board below).
  PAIRED BOARD: uses tiered defense based on non-board high rank.
  [ONE-CARD FLUSH (4+ same suit)]: drop 3.5 levels -> TRASH.
  [ONE-CARD STRAIGHT -- OPEN-ENDED]: drop 2.5 levels -> TRASH.
  [FLUSH POSSIBLE BOARD (3+ same suit)]: drop 1.1(flop) / 0.9(turn) / 0.7(river) levels.
  [STRAIGHT POSSIBLE BOARD -- multiple combos]: drop 0.6(flop) / 0.5(turn) / 0.4(river) levels.
  [STRAIGHT POSSIBLE BOARD -- only 1 combo]: drop 0.4(flop) / 0.3(turn) / 0.2(river) levels.
  NOTE: board texture penalties STACK, EXCEPT: [FLUSH POSSIBLE] and [ONE-CARD FLUSH] do NOT stack (use more severe); [STRAIGHT POSSIBLE] and [ONE-CARD STRAIGHT] do NOT stack (use more severe).
\end{lstlisting}

\medskip\noindent\textbf{Hand class: strong\_draw.}\\
\begin{lstlisting}
Strong draw (combo draw on non-flushy / nut+ flush draw on flushy / flush draw rank>=J on non-flushy / OESD on rainbow non-straighty / combo draw flushy rank>=K):
  ATTACK: 4+ cumulative weighted bets (semi-bluff across multiple streets).
  DEFENSE (by opponent bet sizing, % of pot):
    Flop IP: defend up to 500% pot. Flop OOP: defend up to 400% pot.
    Turn IP: defend up to 190% pot. Turn OOP: defend up to 150% pot.
    Facing all-in: implied odds = ZERO. Need equity >= 60% pot odds to call.
    Facing check-raise: CALL (strong draws have enough equity to continue).
  As aggressor/normal: play aggressively, bet/raise as semi-bluff. Check-raise flop is default.
  As defender: check-call or check-raise depending on equity and position.
  Strategy similar across pot types. Exception: turn SPR <= 1.5 as IP -> check to preserve equity.  COMBO RULE: pair + strong draw -> add ~2.0 extra defense to the PAIR/SHOWDOWN baseline. E.g. second pair (2.5) + strong draw (2.0) = defend ~4.5 cumulative bets.
\end{lstlisting}

\medskip\noindent\textbf{Hand class: medium\_strong\_draw.}\\
\begin{lstlisting}
Medium-strong draw (flush draw rank<J on non-flushy / combo draw flushy rank J-Q / OESD on rainbow+straighty / OESD on two-tone non-straighty (2-card or clean conditions) / K+ flush draw on flushy):
  ATTACK: 3+ cumulative weighted bets (semi-bluff flop + selective turn barrel).
  DEFENSE (by opponent bet sizing, % of pot):
    Flop IP: defend up to 250% pot. Flop OOP: defend up to 200% pot.
    Turn IP: defend up to 100% pot. Turn OOP: defend up to 75% pot.
    Facing all-in: implied odds = ZERO. Need equity >= pot odds to call.
    Flop Facing check-raise: IP: defend up to 150% pot; OOP: defend up to 100% pot.
    Turn Facing check-raise: IP: defend up to 60% pot; OOP: defend up to 40% pot.
  As aggressor: semi-bluff flop, continue turn if equity holds or board improves.
  As normal: bet with high frequency as bluff. Check-raise flop is viable.
  In UNRAISED pots (no postflop raise): consider OOP check-raise as semi-bluff defense.
  COMBO RULE: pair + medium-strong draw -> add ~1.2 extra defense to the PAIR/SHOWDOWN baseline. E.g. second pair (2.5) + medium_strong_draw (1.2) = defend ~3.7 cumulative bets.
\end{lstlisting}

\medskip\noindent\textbf{Hand class: medium\_draw.}\\
\begin{lstlisting}
Medium draw (decent flush draw rank T-Q on flushy board / OESD on two-tone+straighty / 1-card OESD on two-tone non-straighty without clean conditions / combo draw flushy rank 8-T / non-flushy-board gutshot + overcards or backdoor flush on disconnected board):
  ATTACK: 1.5-3 cumulative weighted bets.
  DEFENSE (by opponent bet sizing, % of pot):
    Flop IP: defend up to 150% pot. Flop OOP: defend up to 120% pot.
    Turn IP: defend up to 60% pot. Turn OOP: defend up to 40% pot.
    Facing all-in: implied odds = ZERO. Need equity >= pot odds to call.
    Flop Facing check-raise: IP: defend up to 100% pot; OOP: defend up to 75% pot.
    Turn Facing check-raise: IP: defend up to 40% pot; OOP: defend up to 28% pot.
  As aggressor: double barrel flop+turn if equity holds.
  As normal: bet with high frequency as bluff.
  In UNRAISED pots (no postflop raise): consider OOP check-raise as semi-bluff defense.
  COMBO RULE: pair + medium draw -> add ~0.8 extra defense to the PAIR baseline. E.g. second pair (2.5) + medium draw (0.8) = defend ~3.3 cumulative bets. Do NOT fold pair+draw combos to single bets.
\end{lstlisting}

\medskip\noindent\textbf{Hand class: medium\_weak\_draw.}\\
\begin{lstlisting}
Medium-weak draw (non-flushy 2-card gutshot / 1-card gutshot not at bottom of straight / moderate flush draw rank 6+ on flushy board / combo draw on flushy with high flush cards above / OESD+dual overcards on flushy / dual overcards + double backdoor):
  ATTACK: 1-2 cumulative weighted bets (bluff with equity)
  DEFENSE as % of pot by opponent bet sizing:
    Flop: IP ~94%, OOP ~78%
    Turn: IP ~40%, OOP ~26%
    Facing check-raise: Flop IP ~60%, OOP ~42%; Turn IP ~20%, OOP ~14%
  As aggressor: bet for value+protection on flop, check/fold turn if draw misses
  As defender: call flop bets profitably; turn calls require additional equity
  COMBO RULE: if paired (any pair + medium-weak draw), add ~0.7 extra defense budget
\end{lstlisting}

\medskip\noindent\textbf{Hand class: weak\_draw.}\\
\begin{lstlisting}
Weak draw (bottom gutshot on non-flushy / OESD on flushy without overcards / very small flush draw rank 2-5 on flushy / flushy gutshot + dual overcards / standalone double backdoor draws):
  ATTACK: 0.5-1 cumulative weighted bets (bluff only)
  DEFENSE as % of pot by opponent bet sizing:
    Flop: IP ~68%, OOP ~56%
    Turn: IP ~24%, OOP ~16%
    Facing check-raise: Flop IP ~40%, OOP ~28%; Turn: fold
  As defender: call small flop bets only; fold to turn aggression without improvement
  COMBO RULE: if paired (any pair + weak draw), add ~0.4 extra defense budget
\end{lstlisting}

\medskip\noindent\textbf{Hand class: strong\_overcard\_draw.}\\
\begin{lstlisting}
Strong overcard draw (AK/AQ two premium overcards / KQ or AJ + backdoor straight / KJ/QJ/AT + 2-card backdoor flush or straight / overcard + real draw combo upgrade):
  ATTACK: 1 cumulative weighted bet (bluff with showdown equity)
  DEFENSE as % of pot by opponent bet sizing:
    Flop: IP ~80%, OOP ~65%
    Turn: IP ~35%, OOP ~25%
    Facing check-raise: Flop IP ~55%, OOP ~35%; Turn: fold
  As defender: defend flop at high frequency; turn defense requires improvement potential
  NOTE: texture downgrade may reduce this -- flushy board without flush draw or straighty board reduces tier
  3BP DOWNGRADE: all overcards drop 1 tier except AK, AQ/KQ/AJ+BD flush, AJ/KJ/AT+BD flush+BD straight
  4BP+ DOWNGRADE: all overcards drop 1 tier except AK+BD flush; many weaker overcards become trash
\end{lstlisting}

\medskip\noindent\textbf{Hand class: medium\_overcard\_draw.}\\
\begin{lstlisting}
Medium overcard draw (naked dual overcards with rank sum > 19 like KQ/KJ/QJ/QT/JT/J9 / dual overcards + single backdoor / single overcard T+ with dual backdoor / single overcard T+ with single BD + another card above 2nd board rank / QJ on K-high / QT/JT/Q9 on K-high + BD flush / KQ/KJ on A-high without extras):
  ATTACK: 0.5-1 cumulative weighted bets (bluff only)
  DEFENSE as % of pot by opponent bet sizing:
    Flop: IP ~58%, OOP ~45%
    Turn: IP ~23%, OOP ~15%
    Facing check-raise: Flop IP ~28%, OOP ~18%; Turn: fold
  As defender: call small flop bets selectively; fold turn without improvement
  NOTE: texture downgrade may reduce this -- flushy board without flush draw or straighty board reduces tier
  3BP DOWNGRADE: drops to weak_overcard_draw (most fold to aggression)
  4BP+ DOWNGRADE: drops to weak_overcard_draw or trash depending on hand
\end{lstlisting}

\medskip\noindent\textbf{Hand class: weak\_overcard\_draw.}\\
\begin{lstlisting}
Weak overcard draw (naked dual overcards with rank sum <= 19 like T8/T9/98 / dual overcards + single BD with sum <= 19 / single overcard T+ with BD but no between-pair / JT on Q-high / QT/JT/Q9 on K-high without BD / between-pair single overcard):
  ATTACK: 0.5 cumulative weighted bets (minimal bluff only)
  DEFENSE as % of pot by opponent bet sizing:
    Flop: IP ~35%, OOP ~25%
    Turn: IP ~15%, OOP ~9%
    Facing check-raise: fold
  As defender: fold to most aggression; only call very small bets on flop
  NOTE: texture downgrade may eliminate this entirely -- flushy or straighty board reduces to trash
  3BP DOWNGRADE: drops to trash (no overcard value in 3bp)
  4BP+ DOWNGRADE: drops to trash
\end{lstlisting}

\medskip\noindent\textbf{Hand class: trash.}\\
\begin{lstlisting}
Pure trash (no pair, no overcard, no draw):
  ATTACK in flop/turn: 0-1 (and mostly small sizing stab/c-bet). DEFENSE: 0.
  Zero showdown value. Fold to any bet.
  IP STAB (aggressor checked to you): stab at HIGH frequency.
    Flop/turn: 20-30% pot. River: polarized bluff >60% pot (nothing to lose, max pressure).
  OOP: can make range bets (probe/c-bet at small sizing 20-30% pot). Can make >60% pot bet at river as bluff.
  RIVER BLUFF: If villain checks river to you, consider betting as bluff (zero showdown = nothing to lose). Villain's check signals weakness -- good bluff opportunity.
\end{lstlisting}

\paragraph{Pot-type adjustments.}
The following pot-type context is appended to the hand-class prompt to calibrate budget interpretation:
\begin{lstlisting}
Limp pot: Very wide ranges. Value bet thinner (second pair can be strong). Bottom pair/A-high/K-high hands have more showdown value.

SRP: Standard ranges and thresholds. Top pair can be strong. Bottom pair/A-high can showdown or marginal bluff-catcher. Use standard sizing.

3-bet pot: Both players have NARROWER, STRONGER ranges. Recalibrate hand strength: top pair weak kicker ~ SRP second pair. Overpair below board top card is vulnerable. Two pair+ gains MORE value -- villain pays off with overpairs.

4BP+ pot: Very narrow ranges (premiums only). Range c-bet ~20% pot on most boards. Top pair ~ medium hand (only TPTK+ is strong). Overpair = bread-and-butter value hand. Low SPR -> stacks often committed, play straightforward.
\end{lstlisting}

\paragraph{River bluff and bluff-catch (P5).}
The following supplementary guidance is attached only on the river:
\begin{lstlisting}
=== REFERENCE: River Bluff & Bluff-catch (lower priority -- use as supplementary guidelines) ===
[BLUFFING]
- Blocker: Bluff holding blockers to opponent's value hands (e.g., A spade blocks nut flush).
- No showdown value: Only bluff hands that lose at showdown.
- Consistent story: Bluff line must match strong hand narrative (bet-bet-bet).
- Bluff freq: bet_size/(pot+bet_size) of range should be bluffs.
- Good bluff: blocks opponent's value, doesn't block their bluffs or bluff-catchers.

[BLUFF-CATCHING]
- Need showdown value: Must beat opponent's bluff range (trash still folds).
- Prioritize calling stronger hands (kicker matters)
- Unblocker: Don't block opponent's bluffs (holding draw blockers = bad for catching).
- MDF: Defend >= pot/(pot+bet) to prevent opponent profiting with any bluff.
\end{lstlisting}

\paragraph{Bet-size to weighted-pressure table.}
Weighted pressure is computed from bet size as a percentage of the pot at the time of the bet. The implementation uses a 46-entry piecewise-linear lookup table; representative thresholds are shown below (if bet \% $<$ threshold, the corresponding weight is returned).
\begin{center}\tiny
\setlength{\tabcolsep}{8pt}
\resizebox{\columnwidth}{!}{%
\begin{tabular}{rrrrrrrrrr}
\toprule
Threshold (\% pot) & $<$5 & $<$20 & $<$32 & $<$50 & $<$67 & $<$85 & $<$100 & $<$122 & $<$150 \\
Weight & 0.04 & 0.30 & 0.50 & 0.70 & 0.85 & 1.00 & 1.10 & 1.25 & 1.40 \\
\midrule
Threshold (\% pot) & $<$195 & $<$300 & $<$400 & $<$500 & $<$700 & $<$1000 & $<$1500 & $\geq$1500 & \\
Weight & 1.60 & 2.00 & 2.30 & 2.50 & 2.90 & 3.40 & 4.00 & 4.00 & \\
\bottomrule
\end{tabular}%
}
\end{center}

\paragraph{Special-board override budgets.}
\begin{center}\small
\setlength{\tabcolsep}{10pt}
\begin{tabular}{p{0.18\columnwidth}p{0.8\columnwidth}}
\toprule
Board & Override \\
\midrule
Trips board & Quads/full-house are nuts; nut kicker ATT 0.5 DEF 1.5; second kicker DEF 0.8; lower kickers mostly trash. \\
Double-paired board & Higher-pair full house is nuts; lower full house ATT 2.5 DEF 3.5; flush/straight ATT 2 DEF 3; kicker-only tiers drop in 3BP/4BP. \\
Trips plus side cards & Quads nuts; matching side card or pocket pair forms full-house tiers; flush/straight downgraded below full house. \\
Quads board & Kicker decides: nuts high is nuts, second high ATT 1.5 DEF 2.5, third high ATT 0.5 DEF 1.5. \\
Full-house board & Most hands share board full house; only quads or higher full-house interaction plays aggressively. \\
Board flush & Highest private suited card sets rank: 2nd highest ATT 3 DEF 4 down to no suited card ATT 0 DEF 1.5. \\
Board straight & Usually chopped unless private card improves top end; flush context can dominate the shared straight. \\
\bottomrule
\end{tabular}%
\end{center}

\subsection{Viable Action Computation}

The viable action computation is a complex function of position, street, SPR, role (aggressor or defender), board texture, draw status, and remaining ATT/DEF budget. Rather than a simple lookup table, it uses conditional logic across multiple dimensions. The full implementation is available in the released code. The key factors that determine which actions are shown to the LLM are:

\begin{itemize}
    \item \textbf{Position-specific raise thresholds.} IP and OOP have different raise frequency defaults; OOP check-raises require stronger hands or draws than IP raises.
    \item \textbf{Street-dependent c-bet sizing.} Flop c-bets use smaller sizes on dry boards and larger sizes on wet boards; turn and river bets scale with remaining SPR and board development.
    \item \textbf{Low-SPR commitment logic.} When SPR drops below thresholds (e.g., SPR $\leq$ 1.5), the system commits stacks with hands that have sufficient equity, removing fold from viable options for strong hands and removing raise for marginal hands.
    \item \textbf{Draw-specific defense based on pot odds.} Draw hands use pot-odds thresholds rather than cumulative budget to decide call/fold; the viable action list reflects whether the current bet size is within the draw's defendable range.
    \item \textbf{Role-dependent defaults.} The preflop aggressor defaults to c-betting (bet is shown); the defender defaults to checking (check is shown). These defaults are overridden by budget and board conditions.
    \item \textbf{Paired-board overrides.} On paired and trips boards, the viable action logic adjusts for the increased probability of dominated hands, removing thin value bets that would be profitable on unpaired boards.
\end{itemize}

\section{Evaluation Methodology and AIVAT Variance Reduction}
\label{sec:evaluation-protocol}
\label{sec:aivat-details}

\subsection{AIVAT Overview}

GTOWizard results are reported with AIVAT~\citep{burch2018aivat} in mbb/hand (see Appendix~\ref{sec:hunl-rules} for unit conversions).

AIVAT (Action-Informed Value Assessment Tool) is a variance-reduction technique for poker evaluation. Standard poker evaluation suffers from extremely high variance: a single hand can swing by hundreds of big blinds due to card runout, making raw win-rate estimates unreliable without tens of thousands of hands. AIVAT reduces this variance by leveraging knowledge of the opponent's equilibrium strategy to construct a control variate.

Formally, for each hand $i$, AIVAT computes an adjusted value:
\[
\hat{v}_i^{\text{AIVAT}} = v_i - \sum_{t} \left( \mathbb{E}_{\pi^*}[v | s_t, a_t] - \mathbb{E}_{\pi^*}[v | s_t] \right)
\]
where $v_i$ is the raw outcome, $\pi^*$ is the reference (near-GTO) policy, and the sum runs over decision points where the reference policy's expected values are known. This subtracts the ``luck'' component attributable to card runout while preserving the signal from strategic differences.

The GTOWizard implementation achieves approximately $30\times$ variance reduction compared to raw duplicate outcomes. This means: (i) With AIVAT: $N$ hands yield a standard error equivalent to $30N$ hands of raw play. (ii) Our 5{,}000 hand evaluations achieve the same statistical precision as approximately 150{,}000 hands without AIVAT. (iii) At frontier LLM inference costs (\$70--\$300 per 1{,}000 hands), evaluating 150{,}000 hands would cost \$10{,}500--\$45{,}000 per configuration, which is economically prohibitive for multi-model comparison.

\subsection{Why GTOWizard as Sole Benchmark}

We evaluate exclusively against GTOWizard for three reasons:

\paragraph{1. AIVAT availability.} GTOWizard is the only publicly available HUNL benchmark that provides AIVAT variance-reduced evaluation. Other opponents (Slumbot, rule-based bots, other LLMs) would require raw outcome evaluation, demanding $\sim$30$\times$ more hands for equivalent statistical power.

\paragraph{2. Benchmark strength.} GTOWizard represents current state-of-the-art GTO solving. It beats the 2018 ACPC champion Slumbot by $194 \pm 41$ mbb/hand over 150{,}000 hands. Evaluating against a weaker opponent would be less informative about the ceiling of LLM poker play.

\paragraph{3. Economic constraints.} Frontier LLM inference with extended reasoning is expensive. The total evaluation cost for results reported in this paper exceeds \$2{,}500. Multi-opponent evaluation without AIVAT would multiply this cost by $\sim$30$\times$, making comprehensive ablation studies economically impractical.

\subsection{Experimental Details}

\begin{center}\small

\begin{tabular}{lll}
\toprule
\textbf{LLM Model} & \textbf{Cost/hand} & \textbf{Settings} \\
\midrule
GPT-5.5 XHigh & ${\sim}\$0.30$ & XHigh reasoning, temp 1.0 \\
Claude Opus 4.6 & ${\sim}\$0.07$ & Max thinking (100K tokens), temp 1.0 \\
Claude Opus 4.7 & ${\sim}\$0.07$ & Max thinking (100K tokens), temp 1.0 \\
\bottomrule
\end{tabular}%
\end{center}

All experiments are evaluated for at least 5{,}000 hands against GTOWizard with AIVAT variance reduction. All agents use forced tool-use output (structured JSON via the \texttt{poker\_action} tool schema). Temperature is set to 1.0 as required by the extended-thinking API. No random seeds are applicable: each hand is dealt by the GTOWizard server and plays out deterministically given the model's response. Concurrent workers (6) are used to amortize wall-clock time but do not affect results since hands are independent.

\subsection{Budget Calibration Methodology}
\label{sec:budget-calibration}

The ATT/DEF budget values are designed by human poker experts among the authors, drawing on years of professional play and coaching experience. The calibration process is as follows:

\paragraph{Expert-driven design.} Budget values are set based on the expert authors' deep understanding of poker strategy. For each hand category and board texture, the experts specify how aggressively (ATT) or defensively (DEF) a hand should be played across multiple streets. For example, top pair on a dry board receives ATT $\geq$ 1.0 because expert consensus is that such hands should bet at least one street; marginal hands receive low ATT budgets because experts know they should mostly check. The numeric values are calibrated so that the weighted-pressure cost of standard betting lines (e.g., 33\% flop + 66\% turn + 100\% river) aligns with expert intuition about how many streets each hand class should bet.

\paragraph{Robustness through expert knowledge.} The budget system is designed to be robust across different opponents and game conditions because it encodes \emph{general poker principles} rather than opponent-specific exploits. The experts ensured that:
\begin{itemize}
    \item Board-texture adjustments reflect well-known strategic considerations (e.g., wet boards reduce one-pair hand budgets because equity realization is lower)
    \item Pot-type scaling follows standard theory (3-bet pots require tighter ranges, hence adjusted budgets)
    \item Draw-class thresholds match expert judgment on semi-bluff frequency
\end{itemize}

\paragraph{Generalization evidence.} The same budget tables work across three different LLMs (GPT-5.5, Claude Opus 4.6, and Claude Opus 4.7) with consistent improvement, confirming that the expert-designed values capture general poker knowledge rather than model-specific or opponent-specific patterns.

\section{Error Pattern Analysis}
\label{sec:error-analysis}

Qualitative inspection of high-loss hands reveals three recurring error patterns that persist despite PokerSkill's scaffolding:

\paragraph{Sizing misjudgment.}
The budget correctly permits betting but the LLM selects a suboptimal size. For example, overbetting a medium-strength hand (using a pot-sized bet when a half-pot bet would extract more value from the opponent's calling range) or underbetting a polarized range (using a small bet when the hand's polarity warrants a larger size). This suggests that fine-grained sizing intuition is harder to activate than binary bet/check decisions.

\paragraph{Context boundary ambiguity.}
Hands near the boundary between two classes receive slightly misaligned guidance. For example, a marginal top pair with a weak kicker (5th kicker) may be classified as ``top pair'' and receive the associated budget, when its actual strategic profile is closer to ``second pair'' given the board texture and pot type. This indicates that discrete classification cannot fully capture the continuous nature of hand strength.

\paragraph{Multi-street incoherence.}
Per-decision budgets are locally sound but the three-street sequence is not globally planned. For example, a hand bets the flop (consuming 0.6 budget) and turn (consuming 0.8 budget), then faces a river raise with insufficient remaining budget to continue. The budget system lacks forward-looking search: it cannot anticipate that betting the turn will create an uncomfortable river situation. This failure mode is inherent to per-street scaffolding without global planning.

These observations are qualitative; rigorous decomposition would require a solved equilibrium at full HUNL scale, which is computationally infeasible.



\section{Broader Implications}
\label{sec:broader-impact}

The decision-binding problem, where an LLM possesses relevant knowledge but fails to apply the correct principle at the right moment, could generalize beyond poker. Medical diagnosis requires binding symptoms to the correct differential; legal reasoning requires binding case facts to the relevant statute; negotiation requires binding the current offer to the appropriate concession strategy. In each domain, the challenge is not missing knowledge but \emph{situated selection} among competing heuristics.

PokerSkill demonstrates that structured context analysis combined with selective knowledge retrieval can substantially improve LLM performance without fine-tuning, by performing situation classification deterministically and presenting bounded decisions rather than open-ended generation. The similar architectural pattern (e.g., a deterministic context engine that classifies the situation, a curated knowledge library indexed by situation type, and a bounded decision interface) could be applied to other sequential decision domains where expert knowledge is available but LLMs struggle to activate it reliably.

\subsection{Budget Calibration Methodology}
\label{sec:budget-calibration}

The ATT/DEF budget values are designed by human poker experts among the authors, drawing on years of professional play and coaching experience. The calibration process is as follows:

\paragraph{Expert-driven design.} Budget values are set based on the expert authors' deep understanding of poker strategy. For each hand category and board texture, the experts specify how aggressively (ATT) or defensively (DEF) a hand should be played across multiple streets. For example, top pair on a dry board receives ATT $\geq$ 1.0 because expert consensus is that such hands should bet at least one street; marginal hands receive low ATT budgets because experts know they should mostly check. The numeric values are calibrated so that the weighted-pressure cost of standard betting lines (e.g., 33\% flop + 66\% turn + 100\% river) aligns with expert intuition about how many streets each hand class should bet.

\paragraph{Robustness through expert knowledge.} The budget system is designed to be robust across different opponents and game conditions because it encodes \emph{general poker principles} rather than opponent-specific exploits. The experts ensured that:
\begin{itemize}
    \item Board-texture adjustments reflect well-known strategic considerations (e.g., wet boards reduce one-pair hand budgets because equity realization is lower)
    \item Pot-type scaling follows standard theory (3-bet pots require tighter ranges, hence adjusted budgets)
    \item Draw-class thresholds match expert judgment on semi-bluff frequency
\end{itemize}

\paragraph{Generalization evidence.} The same budget tables work across three different LLMs (GPT-5.5, Claude Opus 4.6, and Claude Opus 4.7) with consistent improvement, confirming that the expert-designed values capture general poker knowledge rather than model-specific or opponent-specific patterns.

\section{Complete Hand Prompt Traces}
\label{sec:hand-traces}

The following example is prompt traces from GPT-5.5 XHigh + PokerSkill. Each listing begins at the preflop decision and continues through the terminal decision, including the original prompt blocks, model responses, validated actions, and environment result lines.


\begin{lstlisting}

============================================================
Street: preflop
Action: b 9.0

--- PROMPT ---
[RANGE TABLE] scenario=bb_vs_open hand=54s

--- GPT RESPONSE ---
{"action": "b", "amount": 9.0, "reasoning": "GTO range table"}

============================================================
Street: flop
Action: b 5.0

--- PROMPT ---
=== HUNL 200BB Hand ===

SITUATION:
- Street: FLOP
- Position: BB (OOP postflop)
- Pot type: 3BP (preflop aggressor: hero)
- You are the AGGRESSOR

YOUR HAND: 5c 4c (54s)
>>> Hand evaluation: STRONG DRAW <<<
BOARD: 7s 6h Jc
Board texture: RAINBOW board; MIXED high/low board

STACKS & POT:
- Pot: 18 | Total pot: 18
- Your stack: 191 | Villain: 191
- SPR: 10.6
- No bet to face

ACTION HISTORY:
  preflop: bet/raise to 2.75 -> bet/raise to 9 -> call

LEGAL ACTIONS: check (k), bet/raise (b) [1 - 191]

=== GENERAL PRINCIPLES (apply in priority order) ===

1. DEFENDER OOP -- CHECK OR DONK: If you are the DEFENDER and OOP, your DEFAULT action is CHECK to the aggressor (let them bet, then check-call or check-raise).
- DONK BET (15% pot) is viable in specific spots: flop low board (high card <= 8, no board pair), or turn/river when a non-top card pairs the board or board becomes double-paired.
- When viable options include DONK BET, weigh it against CHECK -- donk is not mandatory.

2. Flop C-BET: As AGGRESSOR, c-bet is default on most boards.
- High-card / paired boards: c-bet ~25% pot at high frequency.
- Default dry boards: c-bet ~65% pot at moderate frequency.
- Wet / low connected boards (e.g. T98, 753): OOP aggressor -> range check in SRP/3BP. IP aggressor -> lower freq with larger size.
- In 4BP+: range c-bet ~20% pot on almost all boards (your range is much stronger).

3. HAND STRENGTH FIRST: Check [YOUR HAND STRENGTH] below BEFORE applying MDF/pot odds.
- Pure trash (no pair, no draw) -> FOLD to any bet. MDF does NOT apply to unplayable hands.

4. POT CONTROL: Medium-strength hands control pot size.
- IP: check back for pot control (don't bloat pot with marginal hands). If the oop aggressor check to you, you can stab ~25% pot and then check to the river.
- OOP: check is always the mainly option. Sometimes if the IP aggressor check in the previous stage, you can probe/blocking bet ~25% pot for deny equity/river cheap showdown.
- Medium pairs, Small pairs, A-high -> Mostly aim for cheap showdown.

5. ATTACK BUDGET IS BINDING: The >>> ATTACK BUDGET <<< line in SITUATION ANALYSIS is the FINAL authority on whether to bet.
- When it says CHECK -> you MUST check, regardless of board texture, position, or scenario suggestions.
- When it says BET -> betting is allowed (choose sizing from options). Not mandatory.
- ATTACK BUDGET already integrates hand strength, board texture, and cumulative betting history into one decision.
- Do NOT override ATTACK BUDGET with your own judgment about hand strength or board safety.
- EXCEPTION -- RIVER BLUFF: Zero-showdown hands (trash/air) on RIVER may bluff when not facing a bet, regardless of budget. Nothing to lose by bluffing.
- EXCEPTION -- RIVER VALUE: On RIVER, if you are NOT the OOP player acting first (i.e., IP in any case, or OOP facing a bet), value hands with ATTACK remaining >=1 MUST bet or raise. Never flat-call with a value hand on the river unless you are OOP acting first (where check-raise trapping is allowed).
- EXCEPTION -- RIVER SHOWDOWN: On RIVER, if you are NOT OOP acting first, and ATTACK remaining <=0.5, and SPR >0.5: do NOT bet or raise -- check back (IP) or call/fold (facing bet). Thin value on the river risks a raise that costs far more than the thin value gained.

VALUE BETTING: Strong hands bet for value. Size by opponent's calling range.
V:B RATIO: River polarized range -- when betting 1x pot, value:bluff ~ 2:1. Adjust by bet size.
MDF: MDF = pot/(pot+bet). Defend >= this freq. But multi-street / OOP / range disadvantage -> can fold more.
POT ODDS: pot_odds = call/(pot+bet+call). Need equity > pot odds for profitable call.
IMPLIED ODDS: Deep stacks + drawing hands -> effective odds > pot odds. Set mining benefits.
COMMITMENT: SPR <4 -> committed with top pair+ (play straightforward, get stacks in). SPR >10 -> plan 2-3 streets of action. 
BET SIZING:
- Min bet is at least 1BB. Bets can be in decimal form (rounded to 2 decimal places). In small pots (limp pot ~2BB), use 1BB as default bet (~50% pot).
- Flop range bet: ~33% pot (but never below 1BB)
- Turn/river polarized: ~75-100% pot
- Overbet (125-200% pot): polarized bet on nut advantage boards (turn/river only)
- Stab / blocking bet: ~20-35% pot
- MAX: never bet more than 200% of the pot postflop
- GEOMETRIC (PREFERRED): Prefer geometric sizing when it works out to <=150% pot per street. See [GEOMETRIC SIZING] below for exact amount. Use geometric as DEFAULT for value hands planning multi-street betting.
- RIVER ALL-IN: If SPR < 2 on river, consider all-in for both polarized value and bluffs.
- RIVER NO THIN VALUE: On river, do NOT bet < 50% pot for thin value (all-in excepted). Either bet big (>=50% pot) for value, or check back.
- Facing smaller bets -> defend wider. Facing larger bets -> defend tighter.

PAIRED BOARD DEFENSE (flop only, OOP defender facing bet):
- Trips or non-overcard draw -> RAISE SMALL (33% pot). Do NOT flat-call.
- Pair / showdown hands -> normal CALL logic applies.

BET TYPE REFERENCE:
| Type              | Size         | Purpose           |
| POLARIZED BET     | 60-200% pot  | VALUE or BLUFF    |
| MEDIUM BET        | 55-65% pot   | THIN VALUE or BLUFF (river)|
| C-BET             | 15-65% pot   | DENY EQUITY (flop)|
| DELAY C-BET       | 20-35% pot   | DENY EQUITY (turn)|
| STAB              | 20-35% pot   | INFO + DENY       |
| PROBE BET         | 20-35% pot   | INFO + DENY       |
| BLOCK BET         | 20-35% pot   | BLOCK (river)     |
| DONK BET          | 15% pot      | INFO (OOP defend) |
| SEMI-BLUFF LARGE  | 67%+ pot     | BLUFF (draw)      |
| SEMI-BLUFF SMALL  | 20-35% pot   | BLUFF (draw)      |

ATTACK/DEFENSE NOTATION:
- ATTACK N = you can profitably bet/raise a cumulative total of ~N weighted streets (larger bets count more than small bets).
- DEFENSE N = you can profitably call/continue against a cumulative total of ~N weighted bets from opponent. Defend at most N times, at minimum N-1 times.
- RAISE THRESHOLD: when ATTACK >= (number of postflop bets already placed + remaining streets + 1), consider raising when facing opponent's bet, then continue betting on non-dangerous runouts to river. Example: ATTACK>=4 hand on flop in BB facing a c-bet -> consider high-frequency check-raise (1 bet already placed + 2 remaining streets + 1 = 4).
- OOP CHECK-RAISE GUIDANCE (facing bet, non-river): OOP should CHECK-RAISE more than CALL when option available. Vulnerable hands (sets/two-pair on draw-heavy boards) -> HIGH frequency CR to deny equity. Invulnerable hands (dry boards) -> can mix, occasional trap OK.
- IP RAISE vs SMALL BET: IP facing small bets (<=50% pot) should RAISE more often -- don't let opponent see cheap cards. IP facing check-raises or large bets as defender -> mostly CALL to keep ranges capped.
- LOW SPR RAISE: At low SPR, look at ATTACK remaining. Remaining enough -> raise to end hand quickly. Remaining VERY high at low SPR -> can slow-play (trap) since hand is too strong to need protection. If remaining <= 0, do NOT raise for value.
- GEOMETRIC SIZING PRIORITY: As polarized aggressor (NOT c-bet/stab), PREFER geometric sizing from [GEOMETRIC SIZING] section. Geometric gets stacks in efficiently across streets. Use geometric as DEFAULT for value hands; standard sizing (33%, 50%, 75%) only when geometric unavailable or >150% pot.
- CHECK THRESHOLD: when ATTACK < (weighted bets so far + remaining streets + 1), and you can check (not facing a bet), consider CHECKING for pot control. Bet on later streets if opponent shows weakness. Example: ATTACK=2, turn weighted bets=1.35, remaining=1 -> threshold=3.35. ATTACK 2 < 3.35 -> check turn, bet river if villain checks.
- For DRAWS: defense is expressed as max bet sizing (% of pot) you can defend per street, because draw equity depends on pot odds vs bet size.
- COMBO (made hand + draw): when you have BOTH a pair/weak showdown AND a draw, your defense is stronger than either alone. Add ~1 (medium draw), ~0.7 (medium-weak draw), ~0.4 (weak draw), ~0.3 (strong overcard draw), ~0.2 (medium overcard draw), ~0.1 (weak overcard draw) extra defense to the made-hand baseline. On PAIRED BOARD, use the tiered DEFENSE baseline (nut-high 1.5 / second-high 1.0 / third-high 0.8) for COMBO.
- DUAL CLASSIFICATION: when you have BOTH a made hand AND a draw, follow whichever classification gives HIGHER ATTACK/DEFENSE values.
- ALL-IN RULE: when facing all-in or calling means going all-in, implied odds = ZERO. Must have equity >= pot odds to call. Draws lose most value at all-in.
- POSITION ADJUSTMENT: On flop/turn, OOP made hands with SPR > 0.5 -> DEFENSE -0.3 (positional disadvantage costs equity across remaining streets).
- RIVER RE-RAISE PENALTY: On river, facing a raise after your bet (re-raise) -> DEFENSE -0.3 (opponent's raising range on river is extremely strong).

[YOUR HAND STRENGTH]
- Strong draw (combo draw on non-flushy / nut+ flush draw on flushy / flush draw rank>=J on non-flushy / OESD on rainbow non-straighty / combo draw flushy rank>=K):
  ATTACK: 4+ cumulative weighted bets (semi-bluff across multiple streets).
  DEFENSE (by opponent bet sizing, % of pot):
    Flop IP: defend up to 500% pot. Flop OOP: defend up to 400% pot.
    Turn IP: defend up to 190% pot. Turn OOP: defend up to 150% pot.
    Facing all-in: implied odds = ZERO. Need equity >= 60% pot odds to call.
    Facing check-raise: CALL (strong draws have enough equity to continue).
  As aggressor/normal: play aggressively, bet/raise as semi-bluff. Check-raise flop is default.
  As defender: check-call or check-raise depending on equity and position.
  Strategy similar across pot types. Exception: turn SPR <= 1.5 as IP -> check to preserve equity.  COMBO RULE: pair + strong draw -> add ~2.0 extra defense to the PAIR/SHOWDOWN baseline. E.g. second pair (2.5) + strong draw (2.0) = defend ~4.5 cumulative bets.

=== SITUATION ANALYSIS (apply to current spot) ===
[OVERALL]
Before choosing your action, analyze the following about YOUR hand:
1. What is your absolute hand strength? (e.g., top pair good kicker, bottom pair, flush draw)
2. [Flop/Turn Only] What draws do you have? (flush draw, straight draw, backdoor draws, combo draws)
3. What is your relative hand strength? Relative strength depends on: absolute strength, board wetness (wet boards devalue non-nut hands), and opponent's action history (aggressive actions = stronger ranges)
4. What is your plan for future streets/opponent actions? (bet 3 streets, check one street, give up)
5. [Attention] READ the [YOUR HAND STRENGTH] section carefully -- it already classified your hand. Trust it over your own quick judgment.

[POT TYPE ADJUSTMENT]
- 3-bet pot: Both players have NARROWER, STRONGER ranges. Recalibrate hand strength: top pair weak kicker ~ SRP second pair. Overpair below board top card is vulnerable. Two pair+ gains MORE value -- villain pays off with overpairs.

[BOARD TEXTURE -- FLOP]
- Dry/rainbow flop: Aggressor c-bets high freq (~80%). Range advantage matters most. Small sizing (25-33% pot).

[ACTION LINE]
Scenario: F-A2 -- Aggressor OOP. We act first. -> C-bet or check (board-dependent).
Weighted postflop bets so far: 0 (hero: 0). Next bet counts as #1.
-> Hand strength baseline (3BP, 3bp -0.5, 1 weighted bets): second pair+ for value.
-> YOUR HAND: trash + strong draw Cumulative attacks so far: 0.0 weighted bets.
>>> ATTACK BUDGET: 5 (base 0.5, draw floor 5) | used: 0 | remaining: +5.0 | BET or CHECK-RAISE <<<
-> Viable options: C-BET (25% pot)

[BET SIZING REFERENCE (pot = 18.00 BB, use these BB amounts)]
- 20% pot (probe/block) = 3.60 BB
- 25% pot (small bet) = 4.50 BB
- 33% pot (1/3 pot) = 5.94 BB
- 50% pot (half pot) = 9.00 BB
- 66% pot (2/3 pot) = 11.88 BB
- 75% pot (3/4 pot) = 13.50 BB
- 100% pot (pot) = 18.00 BB
- 125% pot (1.25x pot) = 22.50 BB
- 150% pot (1.5x pot) = 27.00 BB
- 200% pot (2x pot) = 36.00 BB
- MAX ALL-IN (capped): 54.00 BB (300% pot cap, stack=191.00)

[GEOMETRIC SIZING]
- Geometric bet: 126% pot = 22.76 BB (SPR=10.6, 3 streets left). RECOMMENDED for multi-street value betting.

[POSITION]
- OOP leading: As aggressor/normal: lead with value + bluffs on favorable boards. As defender: lead only on boards where YOUR range improved more (rare).


What is your action? Respond in JSON: {"action": "X", "amount": N, "reasoning": "..."}

--- GPT RESPONSE ---
{"action":"b","amount":5,"reasoning":"Attack budget allows the listed 25% pot c-bet. With an open-ended straight draw as OOP 3-bet aggressor, small semi-bluff c-betting applies pressure while building equity realization."}

============================================================
Street: turn
Action: k 

--- PROMPT ---
=== HUNL 200BB Hand ===

SITUATION:
- Street: TURN
- Position: BB (OOP postflop)
- Pot type: 3BP (preflop aggressor: hero)
- You are the AGGRESSOR

YOUR HAND: 5c 4c (54s)
>>> Hand evaluation: STRONG DRAW <<<
BOARD: 7s 6h Jc 6c
Board texture: TWO-TONE board (flush draw exists but flush NOT yet possible); PAIRED board (1 pair(s)); MIXED high/low board

STACKS & POT:
- Pot: 28 | Total pot: 28
- Your stack: 186 | Villain: 186
- SPR: 6.6
- No bet to face

ACTION HISTORY:
  preflop: bet/raise to 2.75 -> bet/raise to 9 -> call
  flop: bet/raise to 5 -> call

LEGAL ACTIONS: check (k), bet/raise (b) [1 - 186]

=== GENERAL PRINCIPLES (apply in priority order) ===

1. DEFENDER OOP -- CHECK OR DONK: If you are the DEFENDER and OOP, your DEFAULT action is CHECK to the aggressor (let them bet, then check-call or check-raise).
- DONK BET (15% pot) is viable in specific spots: flop low board (high card <= 8, no board pair), or turn/river when a non-top card pairs the board or board becomes double-paired.
- When viable options include DONK BET, weigh it against CHECK -- donk is not mandatory.

2. Flop C-BET: As AGGRESSOR, c-bet is default on most boards.
- High-card / paired boards: c-bet ~25% pot at high frequency.
- Default dry boards: c-bet ~65% pot at moderate frequency.
- Wet / low connected boards (e.g. T98, 753): OOP aggressor -> range check in SRP/3BP. IP aggressor -> lower freq with larger size.
- In 4BP+: range c-bet ~20% pot on almost all boards (your range is much stronger).

3. HAND STRENGTH FIRST: Check [YOUR HAND STRENGTH] below BEFORE applying MDF/pot odds.
- Pure trash (no pair, no draw) -> FOLD to any bet. MDF does NOT apply to unplayable hands.

4. POT CONTROL: Medium-strength hands control pot size.
- IP: check back for pot control (don't bloat pot with marginal hands). If the oop aggressor check to you, you can stab ~25% pot and then check to the river.
- OOP: check is always the mainly option. Sometimes if the IP aggressor check in the previous stage, you can probe/blocking bet ~25% pot for deny equity/river cheap showdown.
- Medium pairs, Small pairs, A-high -> Mostly aim for cheap showdown.

5. ATTACK BUDGET IS BINDING: The >>> ATTACK BUDGET <<< line in SITUATION ANALYSIS is the FINAL authority on whether to bet.
- When it says CHECK -> you MUST check, regardless of board texture, position, or scenario suggestions.
- When it says BET -> betting is allowed (choose sizing from options). Not mandatory.
- ATTACK BUDGET already integrates hand strength, board texture, and cumulative betting history into one decision.
- Do NOT override ATTACK BUDGET with your own judgment about hand strength or board safety.
- EXCEPTION -- RIVER BLUFF: Zero-showdown hands (trash/air) on RIVER may bluff when not facing a bet, regardless of budget. Nothing to lose by bluffing.
- EXCEPTION -- RIVER VALUE: On RIVER, if you are NOT the OOP player acting first (i.e., IP in any case, or OOP facing a bet), value hands with ATTACK remaining >=1 MUST bet or raise. Never flat-call with a value hand on the river unless you are OOP acting first (where check-raise trapping is allowed).
- EXCEPTION -- RIVER SHOWDOWN: On RIVER, if you are NOT OOP acting first, and ATTACK remaining <=0.5, and SPR >0.5: do NOT bet or raise -- check back (IP) or call/fold (facing bet). Thin value on the river risks a raise that costs far more than the thin value gained.

VALUE BETTING: Strong hands bet for value. Size by opponent's calling range.
V:B RATIO: River polarized range -- when betting 1x pot, value:bluff ~ 2:1. Adjust by bet size.
MDF: MDF = pot/(pot+bet). Defend >= this freq. But multi-street / OOP / range disadvantage -> can fold more.
POT ODDS: pot_odds = call/(pot+bet+call). Need equity > pot odds for profitable call.
IMPLIED ODDS: Deep stacks + drawing hands -> effective odds > pot odds. Set mining benefits.
COMMITMENT: SPR <4 -> committed with top pair+ (play straightforward, get stacks in). SPR >10 -> plan 2-3 streets of action. 
BET SIZING:
- Min bet is at least 1BB. Bets can be in decimal form (rounded to 2 decimal places). In small pots (limp pot ~2BB), use 1BB as default bet (~50% pot).
- Flop range bet: ~33% pot (but never below 1BB)
- Turn/river polarized: ~75-100% pot
- Overbet (125-200% pot): polarized bet on nut advantage boards (turn/river only)
- Stab / blocking bet: ~20-35% pot
- MAX: never bet more than 200% of the pot postflop
- GEOMETRIC (PREFERRED): Prefer geometric sizing when it works out to <=150% pot per street. See [GEOMETRIC SIZING] below for exact amount. Use geometric as DEFAULT for value hands planning multi-street betting.
- RIVER ALL-IN: If SPR < 2 on river, consider all-in for both polarized value and bluffs.
- RIVER NO THIN VALUE: On river, do NOT bet < 50% pot for thin value (all-in excepted). Either bet big (>=50% pot) for value, or check back.
- Facing smaller bets -> defend wider. Facing larger bets -> defend tighter.

PAIRED BOARD DEFENSE (flop only, OOP defender facing bet):
- Trips or non-overcard draw -> RAISE SMALL (33% pot). Do NOT flat-call.
- Pair / showdown hands -> normal CALL logic applies.

BET TYPE REFERENCE:
| Type              | Size         | Purpose           |
| POLARIZED BET     | 60-200% pot  | VALUE or BLUFF    |
| MEDIUM BET        | 55-65% pot   | THIN VALUE or BLUFF (river)|
| C-BET             | 15-65% pot   | DENY EQUITY (flop)|
| DELAY C-BET       | 20-35% pot   | DENY EQUITY (turn)|
| STAB              | 20-35% pot   | INFO + DENY       |
| PROBE BET         | 20-35% pot   | INFO + DENY       |
| BLOCK BET         | 20-35% pot   | BLOCK (river)     |
| DONK BET          | 15% pot      | INFO (OOP defend) |
| SEMI-BLUFF LARGE  | 67%+ pot     | BLUFF (draw)      |
| SEMI-BLUFF SMALL  | 20-35% pot   | BLUFF (draw)      |

ATTACK/DEFENSE NOTATION:
- ATTACK N = you can profitably bet/raise a cumulative total of ~N weighted streets (larger bets count more than small bets).
- DEFENSE N = you can profitably call/continue against a cumulative total of ~N weighted bets from opponent. Defend at most N times, at minimum N-1 times.
- RAISE THRESHOLD: when ATTACK >= (number of postflop bets already placed + remaining streets + 1), consider raising when facing opponent's bet, then continue betting on non-dangerous runouts to river. Example: ATTACK>=4 hand on flop in BB facing a c-bet -> consider high-frequency check-raise (1 bet already placed + 2 remaining streets + 1 = 4).
- OOP CHECK-RAISE GUIDANCE (facing bet, non-river): OOP should CHECK-RAISE more than CALL when option available. Vulnerable hands (sets/two-pair on draw-heavy boards) -> HIGH frequency CR to deny equity. Invulnerable hands (dry boards) -> can mix, occasional trap OK.
- IP RAISE vs SMALL BET: IP facing small bets (<=50% pot) should RAISE more often -- don't let opponent see cheap cards. IP facing check-raises or large bets as defender -> mostly CALL to keep ranges capped.
- LOW SPR RAISE: At low SPR, look at ATTACK remaining. Remaining enough -> raise to end hand quickly. Remaining VERY high at low SPR -> can slow-play (trap) since hand is too strong to need protection. If remaining <= 0, do NOT raise for value.
- GEOMETRIC SIZING PRIORITY: As polarized aggressor (NOT c-bet/stab), PREFER geometric sizing from [GEOMETRIC SIZING] section. Geometric gets stacks in efficiently across streets. Use geometric as DEFAULT for value hands; standard sizing (33%, 50%, 75%) only when geometric unavailable or >150% pot.
- CHECK THRESHOLD: when ATTACK < (weighted bets so far + remaining streets + 1), and you can check (not facing a bet), consider CHECKING for pot control. Bet on later streets if opponent shows weakness. Example: ATTACK=2, turn weighted bets=1.35, remaining=1 -> threshold=3.35. ATTACK 2 < 3.35 -> check turn, bet river if villain checks.
- For DRAWS: defense is expressed as max bet sizing (% of pot) you can defend per street, because draw equity depends on pot odds vs bet size.
- COMBO (made hand + draw): when you have BOTH a pair/weak showdown AND a draw, your defense is stronger than either alone. Add ~1 (medium draw), ~0.7 (medium-weak draw), ~0.4 (weak draw), ~0.3 (strong overcard draw), ~0.2 (medium overcard draw), ~0.1 (weak overcard draw) extra defense to the made-hand baseline. On PAIRED BOARD, use the tiered DEFENSE baseline (nut-high 1.5 / second-high 1.0 / third-high 0.8) for COMBO.
- DUAL CLASSIFICATION: when you have BOTH a made hand AND a draw, follow whichever classification gives HIGHER ATTACK/DEFENSE values.
- ALL-IN RULE: when facing all-in or calling means going all-in, implied odds = ZERO. Must have equity >= pot odds to call. Draws lose most value at all-in.
- POSITION ADJUSTMENT: On flop/turn, OOP made hands with SPR > 0.5 -> DEFENSE -0.3 (positional disadvantage costs equity across remaining streets).
- RIVER RE-RAISE PENALTY: On river, facing a raise after your bet (re-raise) -> DEFENSE -0.3 (opponent's raising range on river is extremely strong).

[YOUR HAND STRENGTH]
- Strong draw (combo draw on non-flushy / nut+ flush draw on flushy / flush draw rank>=J on non-flushy / OESD on rainbow non-straighty / combo draw flushy rank>=K):
  ATTACK: 4+ cumulative weighted bets (semi-bluff across multiple streets).
  DEFENSE (by opponent bet sizing, % of pot):
    Flop IP: defend up to 500% pot. Flop OOP: defend up to 400% pot.
    Turn IP: defend up to 190% pot. Turn OOP: defend up to 150% pot.
    Facing all-in: implied odds = ZERO. Need equity >= 60% pot odds to call.
    Facing check-raise: CALL (strong draws have enough equity to continue).
  As aggressor/normal: play aggressively, bet/raise as semi-bluff. Check-raise flop is default.
  As defender: check-call or check-raise depending on equity and position.
  Strategy similar across pot types. Exception: turn SPR <= 1.5 as IP -> check to preserve equity.  COMBO RULE: pair + strong draw -> add ~2.0 extra defense to the PAIR/SHOWDOWN baseline. E.g. second pair (2.5) + strong draw (2.0) = defend ~4.5 cumulative bets.

=== SITUATION ANALYSIS (apply to current spot) ===
[OVERALL]
Before choosing your action, analyze the following about YOUR hand:
1. What is your absolute hand strength? (e.g., top pair good kicker, bottom pair, flush draw)
2. [Flop/Turn Only] What draws do you have? (flush draw, straight draw, backdoor draws, combo draws)
3. What is your relative hand strength? Relative strength depends on: absolute strength, board wetness (wet boards devalue non-nut hands), and opponent's action history (aggressive actions = stronger ranges)
4. What is your plan for future streets/opponent actions? (bet 3 streets, check one street, give up)
5. [Attention] READ the [YOUR HAND STRENGTH] section carefully -- it already classified your hand. Trust it over your own quick judgment.

[POT TYPE ADJUSTMENT]
- 3-bet pot: Both players have NARROWER, STRONGER ranges. Recalibrate hand strength: top pair weak kicker ~ SRP second pair. Overpair below board top card is vulnerable. Two pair+ gains MORE value -- villain pays off with overpairs.

[BOARD TEXTURE -- TURN]
- Turn pairs the board: As AGGRESSOR: reduce betting frequency on THIS street -- especially when turn pairs a non-top flop card (check at higher frequency). But top pair+ still has 2-3 streets of total value -- if you only bet 1 street so far, you can still bet turn or river. Do NOT assume top pair is now worthless. Trips+ can bet/raise for value. Check-raise from defender is credible.
- Neutral turn (only one flush draw possible, one suit has 2 cards): Only one flush draw is possible -- moderate texture. Made hands retain decent value. Can still bet for value with strong pairs. Check with medium hands if opponent could have the draw.

[ACTION LINE]
Scenario: T-A2 -- We bet flop OOP, villain called. -> Continue or check to defense.
Weighted postflop bets so far: 0.45 (hero: 0.45). Next bet counts as #1.45.
-> Hand strength baseline (3BP, 3bp -0.5, 1.45 weighted bets): second pair+ for value.
-> YOUR HAND: trash + strong draw Cumulative attacks so far: 0.5 weighted bets.
>>> ATTACK BUDGET: 5 (base 0.5, draw floor 5) | used: 0.45 | remaining: +4.5 | BET or CHECK-RAISE <<<
-> Viable options: CHECK

[BET SIZING REFERENCE (pot = 28.00 BB, use these BB amounts)]
- 20% pot (probe/block) = 5.60 BB
- 25% pot (small bet) = 7.00 BB
- 33% pot (1/3 pot) = 9.24 BB
- 50% pot (half pot) = 14.00 BB
- 66% pot (2/3 pot) = 18.48 BB
- 75% pot (3/4 pot) = 21.00 BB
- 100% pot (pot) = 28.00 BB
- 125% pot (1.25x pot) = 35.00 BB
- 150% pot (1.5x pot) = 42.00 BB
- 200% pot (2x pot) = 56.00 BB
- MAX ALL-IN (capped): 84.00 BB (300% pot cap, stack=186.00)

[GEOMETRIC SIZING]
- Geometric bet would be 176% pot (>150%), too large. Use standard sizing instead.

[POSITION]
- OOP leading: As aggressor/normal: lead with value + bluffs on favorable boards. As defender: lead only on boards where YOUR range improved more (rare).


What is your action? Respond in JSON: {"action": "X", "amount": N, "reasoning": "..."}

--- GPT RESPONSE ---
{"action":"k","reasoning":"Viable option is CHECK: the paired turn reduces OOP barreling frequency after villain called flop. With a strong draw, we can realize equity and potentially check-raise if villain bets."}

============================================================
Street: river
Action: b 21.0

--- PROMPT ---
=== HUNL 200BB Hand ===

SITUATION:
- Street: RIVER
- Position: BB (OOP postflop)
- Pot type: 3BP (preflop aggressor: hero)
- NEUTRAL -- aggressor checked prior street, no longer aggressive. You can probe, stab, or bet polarized based on hand strength

YOUR HAND: 5c 4c (54s)
>>> Hand evaluation: TRASH <<<
BOARD: 7s 6h Jc 6c 9h
Board texture: TWO-TONE board (flush draw exists but flush NOT yet possible); STRAIGHT POSSIBLE (2 two-card combos make straight); PAIRED board (1 pair(s)); MIXED high/low board

STACKS & POT:
- Pot: 28 | Total pot: 28
- Your stack: 186 | Villain: 186
- SPR: 6.6
- No bet to face

ACTION HISTORY:
  preflop: bet/raise to 2.75 -> bet/raise to 9 -> call
  flop: bet/raise to 5 -> call
  turn: check -> check

LEGAL ACTIONS: check (k), bet/raise (b) [1 - 186]

=== GENERAL PRINCIPLES (apply in priority order) ===

1. DEFENDER OOP -- CHECK OR DONK: If you are the DEFENDER and OOP, your DEFAULT action is CHECK to the aggressor (let them bet, then check-call or check-raise).
- DONK BET (15% pot) is viable in specific spots: flop low board (high card <= 8, no board pair), or turn/river when a non-top card pairs the board or board becomes double-paired.
- When viable options include DONK BET, weigh it against CHECK -- donk is not mandatory.

2. Flop C-BET: As AGGRESSOR, c-bet is default on most boards.
- High-card / paired boards: c-bet ~25% pot at high frequency.
- Default dry boards: c-bet ~65% pot at moderate frequency.
- Wet / low connected boards (e.g. T98, 753): OOP aggressor -> range check in SRP/3BP. IP aggressor -> lower freq with larger size.
- In 4BP+: range c-bet ~20% pot on almost all boards (your range is much stronger).

3. HAND STRENGTH FIRST: Check [YOUR HAND STRENGTH] below BEFORE applying MDF/pot odds.
- Pure trash (no pair, no draw) -> FOLD to any bet. MDF does NOT apply to unplayable hands.

4. POT CONTROL: Medium-strength hands control pot size.
- IP: check back for pot control (don't bloat pot with marginal hands). If the oop aggressor check to you, you can stab ~25% pot and then check to the river.
- OOP: check is always the mainly option. Sometimes if the IP aggressor check in the previous stage, you can probe/blocking bet ~25% pot for deny equity/river cheap showdown.
- Medium pairs, Small pairs, A-high -> Mostly aim for cheap showdown.

5. ATTACK BUDGET IS BINDING: The >>> ATTACK BUDGET <<< line in SITUATION ANALYSIS is the FINAL authority on whether to bet.
- When it says CHECK -> you MUST check, regardless of board texture, position, or scenario suggestions.
- When it says BET -> betting is allowed (choose sizing from options). Not mandatory.
- ATTACK BUDGET already integrates hand strength, board texture, and cumulative betting history into one decision.
- Do NOT override ATTACK BUDGET with your own judgment about hand strength or board safety.
- EXCEPTION -- RIVER BLUFF: Zero-showdown hands (trash/air) on RIVER may bluff when not facing a bet, regardless of budget. Nothing to lose by bluffing.
- EXCEPTION -- RIVER VALUE: On RIVER, if you are NOT the OOP player acting first (i.e., IP in any case, or OOP facing a bet), value hands with ATTACK remaining >=1 MUST bet or raise. Never flat-call with a value hand on the river unless you are OOP acting first (where check-raise trapping is allowed).
- EXCEPTION -- RIVER SHOWDOWN: On RIVER, if you are NOT OOP acting first, and ATTACK remaining <=0.5, and SPR >0.5: do NOT bet or raise -- check back (IP) or call/fold (facing bet). Thin value on the river risks a raise that costs far more than the thin value gained.

VALUE BETTING: Strong hands bet for value. Size by opponent's calling range.
V:B RATIO: River polarized range -- when betting 1x pot, value:bluff ~ 2:1. Adjust by bet size.
MDF: MDF = pot/(pot+bet). Defend >= this freq. But multi-street / OOP / range disadvantage -> can fold more.
POT ODDS: pot_odds = call/(pot+bet+call). Need equity > pot odds for profitable call.
IMPLIED ODDS: Deep stacks + drawing hands -> effective odds > pot odds. Set mining benefits.
COMMITMENT: SPR <4 -> committed with top pair+ (play straightforward, get stacks in). SPR >10 -> plan 2-3 streets of action. 
BET SIZING:
- Min bet is at least 1BB. Bets can be in decimal form (rounded to 2 decimal places). In small pots (limp pot ~2BB), use 1BB as default bet (~50% pot).
- Flop range bet: ~33% pot (but never below 1BB)
- Turn/river polarized: ~75-100% pot
- Overbet (125-200% pot): polarized bet on nut advantage boards (turn/river only)
- Stab / blocking bet: ~20-35% pot
- MAX: never bet more than 200% of the pot postflop
- GEOMETRIC (PREFERRED): Prefer geometric sizing when it works out to <=150% pot per street. See [GEOMETRIC SIZING] below for exact amount. Use geometric as DEFAULT for value hands planning multi-street betting.
- RIVER ALL-IN: If SPR < 2 on river, consider all-in for both polarized value and bluffs.
- RIVER NO THIN VALUE: On river, do NOT bet < 50% pot for thin value (all-in excepted). Either bet big (>=50% pot) for value, or check back.
- Facing smaller bets -> defend wider. Facing larger bets -> defend tighter.

PAIRED BOARD DEFENSE (flop only, OOP defender facing bet):
- Trips or non-overcard draw -> RAISE SMALL (33% pot). Do NOT flat-call.
- Pair / showdown hands -> normal CALL logic applies.

BET TYPE REFERENCE:
| Type              | Size         | Purpose           |
| POLARIZED BET     | 60-200% pot  | VALUE or BLUFF    |
| MEDIUM BET        | 55-65% pot   | THIN VALUE or BLUFF (river)|
| C-BET             | 15-65% pot   | DENY EQUITY (flop)|
| DELAY C-BET       | 20-35% pot   | DENY EQUITY (turn)|
| STAB              | 20-35% pot   | INFO + DENY       |
| PROBE BET         | 20-35% pot   | INFO + DENY       |
| BLOCK BET         | 20-35% pot   | BLOCK (river)     |
| DONK BET          | 15% pot      | INFO (OOP defend) |
| SEMI-BLUFF LARGE  | 67%+ pot     | BLUFF (draw)      |
| SEMI-BLUFF SMALL  | 20-35% pot   | BLUFF (draw)      |

ATTACK/DEFENSE NOTATION:
- ATTACK N = you can profitably bet/raise a cumulative total of ~N weighted streets (larger bets count more than small bets).
- DEFENSE N = you can profitably call/continue against a cumulative total of ~N weighted bets from opponent. Defend at most N times, at minimum N-1 times.
- RAISE THRESHOLD: when ATTACK >= (number of postflop bets already placed + remaining streets + 1), consider raising when facing opponent's bet, then continue betting on non-dangerous runouts to river. Example: ATTACK>=4 hand on flop in BB facing a c-bet -> consider high-frequency check-raise (1 bet already placed + 2 remaining streets + 1 = 4).
- OOP CHECK-RAISE GUIDANCE (facing bet, non-river): OOP should CHECK-RAISE more than CALL when option available. Vulnerable hands (sets/two-pair on draw-heavy boards) -> HIGH frequency CR to deny equity. Invulnerable hands (dry boards) -> can mix, occasional trap OK.
- IP RAISE vs SMALL BET: IP facing small bets (<=50% pot) should RAISE more often -- don't let opponent see cheap cards. IP facing check-raises or large bets as defender -> mostly CALL to keep ranges capped.
- LOW SPR RAISE: At low SPR, look at ATTACK remaining. Remaining enough -> raise to end hand quickly. Remaining VERY high at low SPR -> can slow-play (trap) since hand is too strong to need protection. If remaining <= 0, do NOT raise for value.
- GEOMETRIC SIZING PRIORITY: As polarized aggressor (NOT c-bet/stab), PREFER geometric sizing from [GEOMETRIC SIZING] section. Geometric gets stacks in efficiently across streets. Use geometric as DEFAULT for value hands; standard sizing (33%, 50%, 75%) only when geometric unavailable or >150% pot.
- CHECK THRESHOLD: when ATTACK < (weighted bets so far + remaining streets + 1), and you can check (not facing a bet), consider CHECKING for pot control. Bet on later streets if opponent shows weakness. Example: ATTACK=2, turn weighted bets=1.35, remaining=1 -> threshold=3.35. ATTACK 2 < 3.35 -> check turn, bet river if villain checks.
- For DRAWS: defense is expressed as max bet sizing (% of pot) you can defend per street, because draw equity depends on pot odds vs bet size.
- COMBO (made hand + draw): when you have BOTH a pair/weak showdown AND a draw, your defense is stronger than either alone. Add ~1 (medium draw), ~0.7 (medium-weak draw), ~0.4 (weak draw), ~0.3 (strong overcard draw), ~0.2 (medium overcard draw), ~0.1 (weak overcard draw) extra defense to the made-hand baseline. On PAIRED BOARD, use the tiered DEFENSE baseline (nut-high 1.5 / second-high 1.0 / third-high 0.8) for COMBO.
- DUAL CLASSIFICATION: when you have BOTH a made hand AND a draw, follow whichever classification gives HIGHER ATTACK/DEFENSE values.
- ALL-IN RULE: when facing all-in or calling means going all-in, implied odds = ZERO. Must have equity >= pot odds to call. Draws lose most value at all-in.
- POSITION ADJUSTMENT: On flop/turn, OOP made hands with SPR > 0.5 -> DEFENSE -0.3 (positional disadvantage costs equity across remaining streets).
- RIVER RE-RAISE PENALTY: On river, facing a raise after your bet (re-raise) -> DEFENSE -0.3 (opponent's raising range on river is extremely strong).

[YOUR HAND STRENGTH]
- Pure trash (no pair, no overcard, no draw):
  ATTACK in flop/turn: 0-1 (and mostly small sizing stab/c-bet). DEFENSE: 0.
  Zero showdown value. Fold to any bet.
  IP STAB (aggressor checked to you): stab at HIGH frequency.
    Flop/turn: 20-30% pot. River: polarized bluff >60% pot (nothing to lose, max pressure).
  OOP: can make range bets (probe/c-bet at small sizing 20-30% pot). Can make >60% pot bet at river as bluff.
  RIVER BLUFF: If villain checks river to you, consider betting as bluff (zero showdown = nothing to lose). Villain's check signals weakness -- good bluff opportunity.

=== SITUATION ANALYSIS (apply to current spot) ===
[OVERALL]
Before choosing your action, analyze the following about YOUR hand:
1. What is your absolute hand strength? (check [YOUR HAND STRENGTH] below -- trust its classification)
2. There are NO draws on the river -- all draws have either completed or missed.
3. What is your relative hand strength? Consider: board texture, opponent's action history across all streets, and what hands opponent is representing.
4. Is this a VALUE BET, BLUFF, or CHECK/CALL situation? Value bet strong hands, bluff with no showdown value + good blockers, check back medium hands.

[POT TYPE ADJUSTMENT]
- 3-bet pot: Both players have NARROWER, STRONGER ranges. Recalibrate hand strength: top pair weak kicker ~ SRP second pair. Overpair below board top card is vulnerable. Two pair+ gains MORE value -- villain pays off with overpairs.

[BOARD TEXTURE -- RIVER]
- Dry river (no flush possible): Board safe -- no draws completed. Made hand values stable. Bluff-catching viable with any showdown value. Bluffs need blockers to be credible.

[ACTION LINE]
Scenario: R-BC2 -- We bet flop OOP, checked turn. -> We can bet polarized or blocking bet or check.
Weighted postflop bets so far: 0.45 (hero: 0.45). Next bet counts as #1.45.
-> Hand strength baseline (3BP, 3bp -0.5, 1.45 weighted bets): second pair+ for value.
-> YOUR HAND: trash Cumulative attacks so far: 0.5 weighted bets.
>>> ATTACK BUDGET: 0.5 (base 0.5) | used: 0.45 | remaining: +0.1 | POLARIZED BET (bluff) or CHECK or BLOCK BET <<<
-> Viable options: POLARIZED BET (bluff) / BLOCK BET (20-35% pot) / CHECK

[BET SIZING REFERENCE (pot = 28.00 BB, use these BB amounts)]
- 66% pot (2/3 pot) = 18.48 BB
- 75% pot (3/4 pot) = 21.00 BB
- 100% pot (pot) = 28.00 BB
- 125% pot (1.25x pot) = 35.00 BB
- 150% pot (1.5x pot) = 42.00 BB
- 200% pot (2x pot) = 56.00 BB
- MAX ALL-IN (capped): 84.00 BB (300% pot cap, stack=186.00)

[GEOMETRIC SIZING]
- Geometric bet would be 664% pot (>150%), too large. Use standard sizing instead.

[POSITION]
- OOP leading: As aggressor/normal: lead with value + bluffs on favorable boards. As defender: lead only on boards where YOUR range improved more (rare).

=== REFERENCE: River Bluff & Bluff-catch (lower priority -- use as supplementary guidelines) ===
[BLUFFING]
- Blocker: Bluff holding blockers to opponent's value hands (e.g., As blocks nut flush).
- No showdown value: Only bluff hands that lose at showdown.
- Consistent story: Bluff line must match strong hand narrative (bet-bet-bet).
- Bluff freq: bet_size/(pot+bet_size) of range should be bluffs.
- Good bluff: blocks opponent's value, doesn't block their bluffs or bluff-catchers.

[BLUFF-CATCHING]
- Need showdown value: Must beat opponent's bluff range (trash still folds).
- Prioritize calling stronger hands (kicker matters)
- Unblocker: Don't block opponent's bluffs (holding draw blockers = bad for catching).
- MDF: Defend >= pot/(pot+bet) to prevent opponent profiting with any bluff.


What is your action? Respond in JSON: {"action": "X", "amount": N, "reasoning": "..."}

--- GPT RESPONSE ---
{"action":"b","amount":21,"reasoning":"With zero showdown value, checking loses almost always, and villain's turn check caps many strong hands. Use a polarized 75% pot bluff rather than a small block bet to pressure pairs and missed floats."}

\end{lstlisting}

\end{document}